\documentclass[lettersize,journal]{IEEEtran}
\usepackage{amsmath,amsfonts}
\usepackage{algorithmic}
\usepackage{algorithm}
\usepackage{array}
\usepackage[caption=false,font=normalsize,labelfont=sf,textfont=sf]{subfig}
\usepackage{textcomp}
\usepackage{stfloats}
\usepackage{url}
\usepackage{multirow}
\usepackage{verbatim}
\usepackage{graphicx}
\usepackage{cite}
\usepackage{booktabs}
\usepackage[table,xcdraw]{xcolor}
\begin{document}

\title{DIP: Diffusion Learning of Inconsistency Pattern for General DeepFake Detection}

\author{Fan Nie, Jiangqun Ni, ~\IEEEmembership{Member,~IEEE}, Jian Zhang, Bin Zhang, Weizhe Zhang, ~\IEEEmembership{Senior Member,~IEEE,}

\thanks{Manuscript received April 15, 2024; revised July 8, 2024; accepted September 8, 2024. This work was supported in part by the National Natural Science Foundation of China under Grants
U23B2022 and U22A2030; in part by Guangdong Major Project of Basic and Applied Basic Research under Grant 2023B0303000010; in part by the Major Key Project of PCL under Grant PCL2023A05. The associate
editor coordinating the review of this article and approving it for publication
was Dr. Richang Hong. (\emph{Corresponding author: Jiangqun Ni.})}

\thanks{Fan Nie is with the School of Computer Science and Engineering, Sun Yat-sen University, Guangzhou 510006, China, and also with the Department of New Networks, Pengcheng Laboratory, Shenzhen 518066, China (e-mail: nief6@mail2.sysu.edu.cn).}
\thanks{Jiangqun Ni is with the School of Cyber Science and Technology, Sun Yat-sen University, Shenzhen 510275, China, and also with the Department of Networks, Pengcheng Laboratory, Shenzhen 518066, China (e-mail: issjqni@mail.sysu.edu.cn).}
\thanks{Jian Zhang is with the School of Computer Science and Engineering, Sun Yat-sen University, Guangzhou 510006, China (e-mail: zhangj266@mail2.sysu.edu.cn).}
\thanks{Bin Zhang is with the Department of Networks, Pengcheng Laboratory, Shenzhen 518066, China (e-mail: bin.zhang@pcl.ac.cn).}
\thanks{Weizhe Zhang is with the School of Cyberspace Science, Harbin Institute
of Technology, Harbin 150001, China, and also with the Department of New Networks, Peng Cheng Laboratory, Shenzhen 518066, China (e-mail: wzzhang@hit.edu.cn).}
}


\markboth{IEEE TRANSACTIONS ON MULTIMEDIA,~Vol.~xx, No.~x, August~2024}%
{NIE \MakeLowercase{\textit{et al.}}: DIP: Diffusion Learning of Inconsistency Pattern for General DeepFake Detection}

\IEEEpubid{0000--0000/00\$00.00~\copyright~2021 IEEE}

\maketitle

\begin{abstract}
With the advancement of deepfake generation techniques, the importance of deepfake detection in protecting multimedia content integrity has become increasingly obvious. Recently, temporal inconsistency clues have been explored to improve the generalizability of deepfake video detection. According to our observation, the temporal artifacts of forged videos in terms of motion information usually exhibits quite distinct inconsistency patterns along horizontal and vertical directions, which could be leveraged to improve the generalizability of detectors. In this paper, a transformer-based framework for \textbf{D}iffusion Learning of \textbf{I}nconsistency \textbf{P}attern (DIP) is proposed, which exploits directional inconsistencies for deepfake video detection. Specifically, DIP begins with a spatiotemporal encoder to represent spatiotemporal information. A directional inconsistency decoder is adopted accordingly, where direction-aware attention and inconsistency diffusion are incorporated to explore potential inconsistency patterns and jointly learn the inherent relationships. In addition, the SpatioTemporal Invariant Loss (STI Loss) is introduced to contrast spatiotemporally augmented sample pairs and prevent the model from overfitting nonessential forgery artifacts. Extensive experiments on several public datasets demonstrate that our method could effectively identify directional forgery clues and achieve state-of-the-art performance.
\end{abstract}


\section{Introduction}
\IEEEPARstart{W}{ith} the rapid development of AI generative models and social networks, a large number of fake videos generated by advanced face forgery methods such as Deepfake \cite{noauthor_deepfakes_2024}, Face2Face \cite{thies_face2face_2016}, Faceswap \cite{marek_faceswap_2020}, HifaFace \cite{wang_high-fidelity_2022}, and NeuralTexture \cite{thies_deferred_2019} could be publicly disseminated, which could disrupt the order of cyberspace, and raise serious security concerns. To ensure the authenticity and integrity of multimedia data, it is of great significance to develop effective deepfake detection methods.

\begin{figure}[t]
    \centering
    \includegraphics[width=0.7\linewidth]{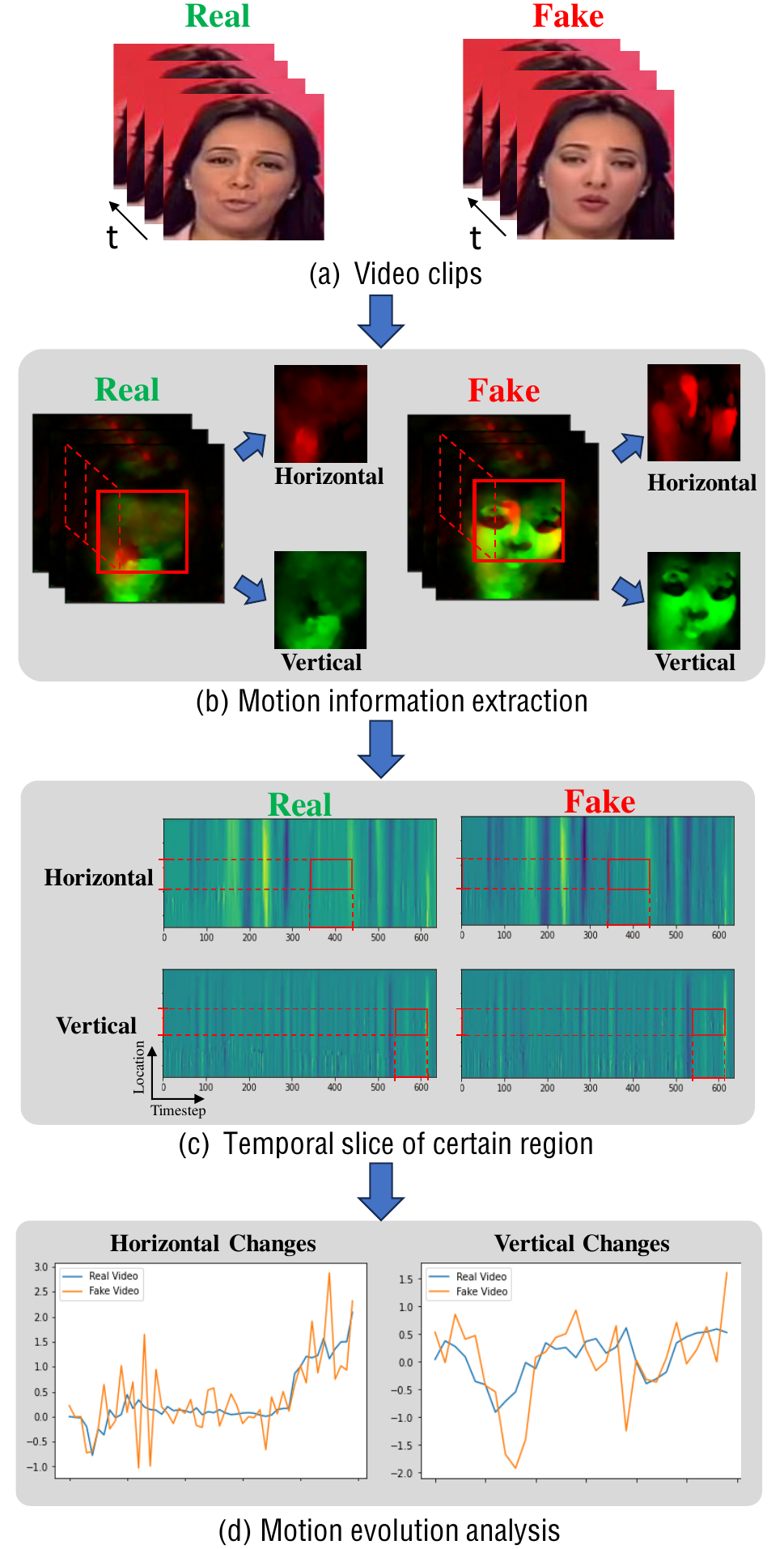}
    \caption{Illustration of the temporal inconsistencies. For a pair of real and fake videos, the motion information in terms of optical flow is extracted and visualized with the TVL1 algorithm \cite{wedel_improved_2009}. Each optical flow frame is then sliced to obtain horizontal and vertical motion slices for real and fake videos. The comparison between real and fake videos for the average temporal motion evolution reveals the inconsistency along both the horizontal and vertical directions.}
    \label{fig:motivation}
\end{figure}

\IEEEpubidadjcol
Prior studies \cite{afchar_mesonet_2018, rossler_faceforensics_2019, amerini_deepfake_2019} formulate deepfake detection as a vanilla classification task and achieve satisfactory performance for the in-dataset scenario. However, their performance in terms of generalizability usually drops significantly under cross-dataset evaluation, where unseen facial manipulation techniques, data distributions, and distortions are involved. This prevents the detectors from being deployed in real-world applications. Recently, the spatiotemporal inconsistencies arising from the facial manipulation have been explored \cite{gu_spatiotemporal_2021, zhang_detecting_2021, gu_region-aware_2022, guan_delving_2022} to capture forgery clues, exhibiting promising generalizability. Zhang et al. \cite{zhang_detecting_2021} utilized convolutional neural networks (CNNs) and temporal dropout to extract discriminative clues. Gu et al. \cite{gu_region-aware_2022} proposed a region-aware temporal filter deployed in CNNs to capture local spatiotemporal inconsistencies. Similarly, Guan et al. \cite{guan_delving_2022} incorporated the advanced Vision Transformer \cite{dosovitskiy_image_2021} and used a local sequence transformer to capture long-term temporal inconsistencies. In general, both spatial and temporal artifacts could be explored to expose deepfake videos. However, detectors without elaborate spatiotemporal modeling tend to capture more salient clues in either the spatial or temporal domain \cite{wang_altfreezing_2023} and do not take full advantage of spatiotemporal representation, leading to less generalization capability. Therefore, how to effectively represent the spatial and temporal forgery cues as a whole remains an open issue in the image forensic community.

To address the aforementioned problem, we extract and visualize the motion information in terms of optical flow for a pair of real and fake video clips with the TVL1 algorithm \cite{wedel_improved_2009}, as shown in Fig. \ref{fig:motivation}(b), where red and green correspond to the displacements along the horizontal and vertical directions and brighter colors indicate larger values. We then slice each optical flow frame vertically at a fixed horizontal location, i.e., within the red dotted box in Fig. \ref{fig:motivation}(b), and concatenate them in terms of horizontal and vertical displacements to obtain both the horizontal and vertical motion slices for real and fake videos respectively, as shown in Fig. \ref{fig:motivation}(c). Fig. \ref{fig:motivation}(d) shows the comparison between real and fake videos for the average temporal motion evolution within the solid red box in Fig. \ref{fig:motivation}(c) along the horizontal and vertical directions. It is quite evident that the motion slices for real video (either horizontal or vertical) evolve much more smoothly whereas those for fake video exhibit more discontinuous patterns. It is the direction of motion that allows us to characterize the temporal inconsistency from a new perspective. In addition, a close observation of Fig. \ref{fig:motivation} reveals that the temporal motion for both real and fake videos features local similarity and diffusion. Specifically, in an optical flow frame, the regions, that are spatially close to each other, exhibit similar motion patterns (local similarity). The motion patterns in one region would gradually affect its surroundings to some extent, known as diffusion, which has been mainly ignored by prior studies.

Based on the above observations, we propose the Diffusion Learning of Inconsistency Pattern (DIP) framework with a vision transformer. For the backbone of DIP, considering the difference in representation between spatial and temporal forgery artifacts, the asymmetric spatiotemporal attention mechanism is adopted to balance the interaction between spatial and temporal artifacts. To effectively capture the forgery clues along horizontal and vertical directions, directional embeddings are obtained through directional pooling operations. The DIP structure allows us to take the temporal inconsistency and the characteristics of temporal motion for both real and fake videos into consideration and to represent the forgery videos through the following modules:

\begin{itemize}
    \item \textbf{Inconsistency Diffusion Module (IDM).} As shown in Fig. \ref{fig:motivation}(b), the temporal motions of both real and fake videos exhibit diffusion effects along the horizontal and vertical directions, and have their own diffusion patterns. The term  "diffusion" is used to illustrate the extent to which the motion of one region affects its surroundings. The IDM is introduced to learn the diffusion intensity (distance) \cite{farbman_diffusion_2010, sun_neural_2019} among neighboring regions along the horizontal and vertical directions, where the IDM regards directional region features as graph nodes and incorporates graph-based diffusion to learn regional diffusion intensities effectively. Equipped with this module, the DIP could effectively capture the forgery artifacts in the horizontal and vertical directions from the perspective of inconsistency diffusion. On the other hand, the learned diffusion patterns (distances) are exploited to optimize the spatiotemporal features with the DIP backbone in a weakly supervised manner.

    \item \textbf{Directional Cross Attention (DiCA).} With DiCA, direction-aware multi-head cross attention is adopted to learn general forgery features through directional interaction. By cross-attention between the horizontal and vertical directions, DiCA could effectively characterize the directional discrepancies between real and fake videos and learn more discriminative and general forgery features for deepfake detection. In addition, the diffusion patterns obtained by IDM are incorporated in DiCA as attention bias to provide a comprehensive view of directional interactions.
\end{itemize}

To further improve the generalizability and robustness performance of the proposed DIP, spatiotemporal data augmentation (DA) is implemented to capture critical information in facial forgeries for deepfake detection. Unlike the work in \cite{zhang_detecting_2021, wang_altfreezing_2023}, both spatial and temporal DAs are adopted by the well-devised triplet data structure, i.e., anchor, positive and negative, to suppress specific and trivial forgery traces. For spatial DA, the specific spatial clues, e.g., high-frequency artifacts, color mismatches, and noise patterns, are effectively mitigated by applying Gaussian blur, color saturation, Gaussian noise, etc., whereas for temporal DA, frame dropping and repeating are employed to attenuate the temporal motion artifacts. The proposed spatiotemporal DA facilitates the model to learn more general forgery representations and refrain it from getting trapped in the local optima. On the other hand, a SpatioTemporal Invariant Loss (STI Loss) is developed to incorporate the spatiotemporally augmented samples in a triplet, which pulls the anchor and positive samples closer while pushing the negative samples from the anchor in the feature space, thus improving the generalizability of the DIP.

The main contributions of the paper are summarized as follows:
\begin{itemize}
    \item A new deepfake video detection framework - DIP is proposed, which exploits the spatiotemporal inconsistency of deepfake video along the horizontal and vertical directions for general deepfake detection.

    \item The Directional Cross Attention (DiCA) is devised to model the temporal inconsistency of deepfake video along horizontal and vertical directions, while the Inconsistency Diffusion Module (IDM) is designed to characterize the temporal artifacts in terms of diffusion patterns along horizontal and vertical directions. These two modules facilitate the model in learning more discriminative spatiotemporal representations for deepfake detection.
    
    \item A new SpatioTemporal Invariant Loss (STI Loss) is developed, which incorporates spatiotemporal data augmentation with triplet structure to drive the model to learn prominent representations for general and robust deepfake detection.
    
    \item Experimental results demonstrate the effectiveness of the proposed modules and the superior performance in terms of generalizability and robustness.
\end{itemize}

\section{Related Work}
\subsection{DeepFake Detection}
Owing to the diversity of practical applications and the agnostic attacks of underlying channels, performance in terms of generalizability and robustness is the primary goal of deepfake detection. To this end, the following inconsistency clues are explored for deepfake video detection, i.e., spatial frequency clues, postprocessing clues, and temporal clues. In \cite{zhao_multi-attentional_2021, zhang_patch_2022}, spatial clues are captured by integrating local artifacts with high-level semantic features. Equipped with DCT and Wavelet transform, \cite{chen_local_2021, liu_spatial-phase_2021, li_frequency-aware_2021, yu_detection_2022} exploit both spatial and frequency inconsistency learning to enhance the generalization performance. On the other hand, the artifacts that appear in facial boundaries due to the blending operations of existing deepfake techniques are also leveraged to significantly improve the performance of deepfake detection \cite{li_face_2020, zhao_learning_2021, shiohara_detecting_2022}.

Compared with image-based methods, video-based methods \cite{gu_region-aware_2022, zheng_exploring_2021, haliassos_lips_2021, khan_video_2021, sun_improving_2021} take forgery clues in the temporal domain into consideration, e.g., the visual and biological inconsistency patterns such as abnormal facial movements \cite{sun_improving_2021, khan_video_2021}, inconsistent eye blinking \cite{li_ictu_2018}, non-synchronous lip movements \cite{haliassos_lips_2021}. With the development of more advanced techniques, the generated facial images tend to be more visually convincing, which would inevitably disable the above methods. 

Given all this, in this paper, the spatiotemporal inconsistencies are characterized with directional temporal representations (along the horizontal and vertical directions) via directional interaction and diffusion learning for general deepfake detection. 

\subsection{Temporal Analysis}
Many efforts have been made with the hope of modeling temporal dependency, which is the essence of video-relevant tasks. Initially, 3D CNNs are usually used for temporal modeling, although they are computationally intensive. Some methods \cite{karpathy_large-scale_2014, simonyan_two-stream_2014} incorporate 2D CNNs with different dimensions to perform temporal modeling, which, however, fails to capture the long-term dependency among video frames. Owing to its prominent long-term modeling capability, the transformer-based methods \cite{bertasius_is_2021, arnab_vivit_2021} are developed for temporal modeling and achieve promising results.

Recently, the transformer-based architecture has also been adopted to learn the spatiotemporal inconsistencies for deepfake video detection. In \cite{khan_video_2021, zheng_exploring_2021, zhao_istvt_2023}, stacked CNNs and transformers are used to learn frame-level forgery representations and expose the temporal forgery clues by exploring the multi-frame forgery representation. In \cite{guan_delving_2022}, the hybrid model of 3D CNN and transformer are developed to learn the temporal inconsistency of deepfake videos. However, its performance for stationary videos is less prominent due to not taking full consideration of the spatial clues.

For the proposed DIP, both the spatial and temporal inconsistencies are taken into consideration by transforming the input video clip into directional representations. The inconsistency between real and fake videos is then modeled from two different perspectives, i.e., directional cross-attention and diffusion, which contributes to better spatiotemporal inconsistency learning.

\begin{figure*}[]
    \centering
    \includegraphics[width=\textwidth]{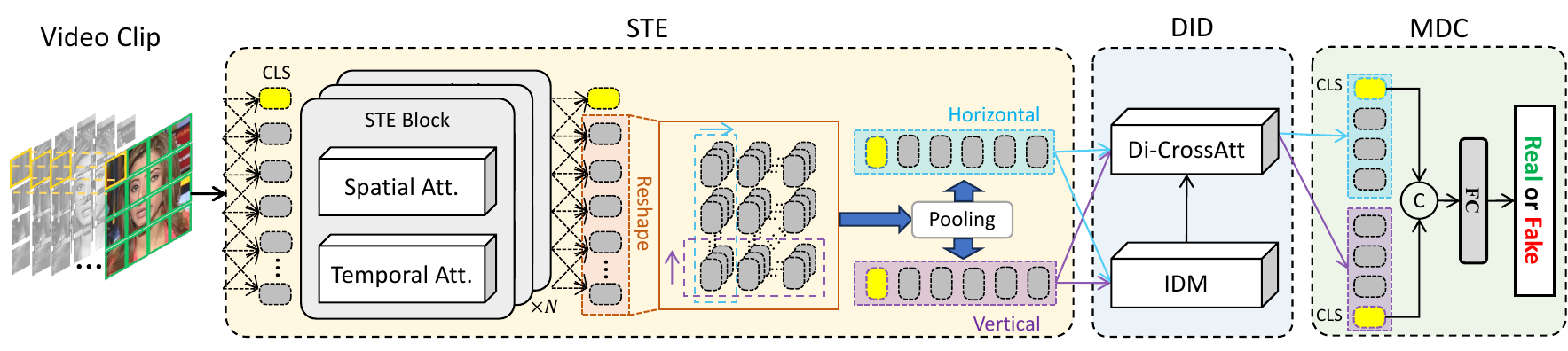}
    \caption{Overview of the proposed DIP. STE extracts forgery spatiotemporal features with embedded sequences. The DID then models inconsistency patterns and fuses features, and MDC exploits the classification tokens for final prediction.}
    \label{fig:overview}
\end{figure*}
\section{Method}
\subsection{DIP Pipeline}
As illustrated in Fig. \ref{fig:overview}, the proposed DIP framework mainly consists of three components, i.e., a SpatioTemporal Encoder (STE), a Joint Directional Inconsistency Decoder (DID), and a Cross-Direction Classifier (MDC), whose functionalities are as follows:
\begin{itemize}
    \item The STE first extracts spatiotemporal features with a unified transformer structure. The extracted features are then split into two directional features via directional pooling operations, which serve as the key assets for pattern modeling and feature fusion.
    \item  The DID is adopted to learn directional inconsistency patterns simultaneously. Equipped with Directional Cross Attention (DiCA) and Inconsistency Diffusion Module (IDM), the DID is expected to learn a better cross-direction inconsistency representation for deepfake detection.
    \item The well-learned directional inconsistency features are then exploited by MDC to classify fake and real videos.
\end{itemize}
\subsection{SpatioTemporal Encoder}
Given the input video clip $V \in \mathbb{R}^{T \times M \times M \times 3}$ with $T$ RGB frames, and without loss of generality, the square frame of size $M \times M$ is assumed, each frame is decomposed into $L \times L$ non-overlapping patches of size $P \times P, L= M /P$ as shown in Fig. \ref{fig:overview}. The video clip $V$ is then transformed into the spatiotemporal token sequence $X \in \mathbb{R}^{T \times L \times L \times E}$ by linearly embedding each patch in $V$ into a $E$-dim token. Let $X(t,i,j)$ denotes the $(i,j)$-th token in $t$-th frame, we then have $\hat{X} \in \mathbb{R}^{T \times (L \cdot L + 1) \times E}$ by appending a classification token in the first position of each token frame $\{X(t,0,0)\}, t=1, \cdots, T$, and embedding the spatial relation $e_{s} \in \mathbb{R}^{(L \cdot L + 1) \times E}$ and the temporal relation $e_{t} \in \mathbb{R}^{T \times E}$ into each token $X(t,i,j), t= 1, \cdots, T,$ and $ i,j = 1, \cdots, L$.

\begin{figure}[h]
    \centering
    \includegraphics[width=\linewidth]{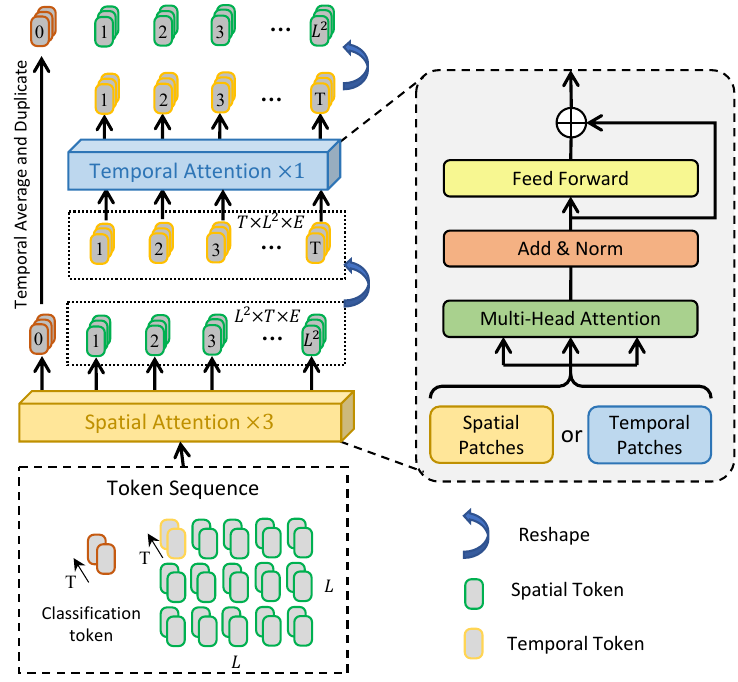}
    \caption{Illustration of a unit STE block. Spatial attention is used to extract spatial dependency for each frame, and temporal attention is applied to characterize the temporal dependency at a specific location across multiple frames.}
    \label{fig:ste_block}
\end{figure}

The token sequence $\hat{X}$ is then fed into the SpatioTemporal Encoder (STE) to obtain the discriminative spatiotemporal representation $Z \in \mathbb{R}^{T \times (L \cdot L + 1) \times E}$, which consists of three stacked asymmetric spatiotemporal attention modules, each implementing three-layer successive spatial attentions for each token frame $\{\hat{X}(t,i,j)\}_{i,j}$ of size $L \times L$ followed by one-layer temporal attention for each temporal token sequence $\{\hat{X}(t,i,j)\}_{t}$ of size $T$ at spatial position $(i,j)$ as shown in Fig. \ref{fig:ste_block}. Note that we exclude classification tokens when forwarding token sequence into temporal attention. To better explore the fine-grained directional forgery clues, the spatiotemporal representation $Z$ is then transformed int both horizontal and vertical sequences $Z_{h}$ and $Z_{v}$ of size $L \times E$ via spatial and temporal pooling operations. For $Z_{h}$, we first perform spatial pooling by averaging or maximizing each row $\{\hat{X}(t,i,j)\}_{j}$ ($i$-th row) in $t$-th frame to calculate the token sequence $\{\hat{Z}(t,i)\}_{i}$ of size $L$, which is then averaged or maximized over frames for temporal pooling to obtain $Z_{h}$ as illustrate Fig. \ref{fig:overview}. And the vertical sequence $Z_{v}$ could be derived similarly.

\subsection{Joint Directional Inconsistency Decoder}
With the directional spatiotemporal representation in terms of $Z_{h}$ and $Z_{v}$, we then proceed to develop the Joint Directional Inconsistency Decoder (DID) to obtain the directional inconsistency representation. The DID is used to integrate the directional features by incorporating the inconsistency patterns. According to the temporal characteristics of real/fake videos, as shown in Fig. \ref{fig:motivation}, the inconsistency patterns are characterized from two different perspectives, i.e., directional interaction and diffusion, which are obtained by the proposed Directional Cross Attention (DiCA) and Inconsistency Diffusion Module (IDM). In addition to the attention scores calculated by DiCA, the diffusion distances by IDM are also taken advantage of to learn the inherent inconsistency relationships among directional features.
\subsubsection{Inconsistency Diffusion Module (IDM)}
As illustrated in Fig. \ref{fig:motivation}, there are quite distinct motion diffusion differences between real and fake videos, which have not been explored in prior arts.  Modeling such diffusion patterns provides a different view to characterize the crucial distinction between real and fake videos. To this end, a graph-based Inconsistency Diffusion Module (IDM) is adopted to calculate the diffusion distances \cite{coifman_diffusion_2006} among nodes on the graph structure via multistep random walking. With the horizontal and vertical sequences $Z_{h}$, $Z_{v}$, the IDM is used to characterize not only the diffusion patterns along the same direction ($Z_{h}$ or $Z_{v}$) but also the ones cross the directions (between $Z_{h}$ and $Z_{v}$).

We exclude classification tokens of both sequences and construct a graph $G = (S, \mathcal{E})$ with node set $S$ of size $2L$, comprising two types of nodes, i.e., $S^{h} \in \{s^{h}_{1},s^{h}_{2},\cdots,s^{h}_{L}\}$ from horizontal token sequence and $ S^{v} \in \{s^{v}_{1},s^{v}_{2},\cdots,s^{v}_{L}\}$ from vertical token sequence, and edge set $\mathcal{E}$. Note that $f^{h}_{i}$ and $f^{v}_{i} (i = 1, \cdots, L)$ represent the embeddings for the node $s^{h}_{i}$ and $s^{v}_{i}$ respectively, and edge set $\mathcal{E}$ could be derived from the associated nodes and their neighborhoods.

\emph{Computation of the Transition Matrix.} For the graph $G = (S, \mathcal{E})$, we assume that any two nodes in $G$ with closer spatiotemporal embeddings are more similar to each other in motion pattern, and the transition matrix among nodes is then used to illustrate the motion patterns for real and fake videos, where two similar nodes have greater transition probabilities. Note that the transition matrix serves to measure two types of similarities among nodes in $G$, i.e., the similarity between nodes in the same direction ($Z_{h}$ or $Z_{v}$) and the one cross the direction ($Z_{h}$ and $Z_{v}$).

We then define the score matrix $W$ among nodes in $G$, comprising four submatrices, i.e.,

\begin{equation}
W = 
\begin{bmatrix}
    W_{hh} & W_{hv} \\
    W_{vh} & W_{vv} \\
\end{bmatrix},
\end{equation}
where $W_{hh}$ and $W_{vv}$ are used to measure the similarity between two nodes in the same direction, while $W_{hv}$ and $W_{vh}$ represent the similarity between two nodes from a different direction, and each of size $L \times L$. For $W_{hh}$, let $i$ and $j \in N_{h}^{h}(i)$ be the node and its neighbor in $S^{h}$, we have

\begin{equation} 
    W_{hh}(i,j) = 
    \begin{cases}
        \mathrm{exp}(-\mu \vert\vert f^{h}_{i}-f^{h}_{j} \vert\vert_{2}^2), & \mathrm{for} \quad j \in N_{h}^{h} (i), \\
        0, & otherwise,
    \end{cases}
\label{eqution: neighbor}
\end{equation}
where $\mu$ is a learnable parameter that is initialized as 0.05, and neighbor size $k_{n} = |N_{h}^{h}(i)|$ is a hyperparameter. For constructing cross-direction submatrices, e.g., $W_{hv}$, we define the cross-direction neighbor set $N^{h}_{v}(i)$ of the node $i \in S^{h}$, which includes node $s^{v}_{j} \in S^{v}, j \in \{i- \left\lfloor\frac{k_{n}}{2}\right\rfloor, \cdots, i + \left\lfloor\frac{k_{n}}{2}\right\rfloor \}$. And $W_{vh}$  could be obtained similarly.

\begin{figure}
    \centering
    \includegraphics[width=\linewidth]{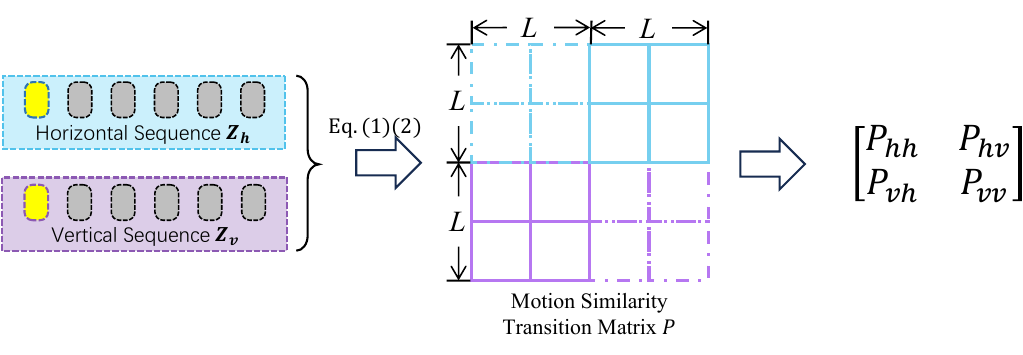}
    \caption{Calculation of motion similarity transition matrix with horizontal and vertical token sequences zh and zv.  The transition matrix $P$ of $2L \times 2L$ consists of four types of submatrices $(L \times L)$, i.e., horizontal-horizontal transition (blue dotted, $P_{hh}$), horizontal-vertical transition (blue solid, $P_{hv}$), vertical-horizontal transition (purple solid, $P_{vh}$), and vertical-vertical transition (purple dotted, $P_{vv}$).}
    \label{fig:inconsistency_matrix}
\end{figure}

Note that, according to our arrangement for score matrix $W$, for graph $G$ with $2L$ nodes, the node $s_{i} \in S^{h}, i \in [1, \cdots, L]$, while $s_{i} \in S^{v}, i \in [L+1, \cdots, 2L]$. We then have the probabilistic transition matrix $P$:

\begin{equation}
    P = D^{-1}W,
\end{equation}
where $D$ is a diagonal normalization matrix $D(i,i) = \sum_{j}W(i,j)$, and $P = [ p(i,j)]$ denotes the probability of stochastic walking from $s_{i}$ to $s_{j}$ in one step. As illustrated in Fig. \ref{fig:inconsistency_matrix}, the transition matrix $P$ is symmetric and could be used to characterize spatiotemporal motion changes or motion diffusion patterns in both directions.

\emph{Motion Similarity Diffusion.} With the transition matrix $P = [p(i,j)]$, we have $P^{t}[p^{t}(i,j)] = \overbrace{P \cdot P \cdots P}^{t\enspace \mathrm{times}}$, where $p^{t}(i,j)$ represents the transition probability through $S$ in $t$ steps from $s_{i}$ to $s_{j}$. Following \cite{sun_neural_2019}, the motion similarity diffusion distance between $s_{i}$ and $s_{j}$ could be defined as the sum of the squared difference between the probabilities \textcolor{red}{of} a random walker starting from $s_{i}$ and $s_{j}$, and ending up at the same node in graph $G$ in $t$ steps, i.e.,
\begin{equation}
    D^{t}(i,j) = \sum_{k}(P^{t}(i,k) - P^{t}(j,k))^{2} \hat{w}(k),
    \label{eqution: diffu}
\end{equation}
where $\hat{w}(k)$ is the reciprocal of the local density at $s_{k}$. The diffusion distance $D^{t}(i,j)$ is smaller if nodes $s_{i},s_{j}$ are connected with intermediate nodes with similar transition probabilities, indicating that the two nodes have a similar motion diffusion pattern.

\emph{Computational Acceleration for Diffusion.}
In practice, iterative computation of Eq. (\ref{eqution: diffu}) is computationally intensive, and the diffusion process could be significantly accelerated by spectral decomposition of $D^{t}$. According to \cite{farbman_diffusion_2010}, $D^{t}$ has a set of $2L$ real eigenvalues $\{ \lambda_{r}\}^{2L}_{1}$: $\lambda_{1} = 1 \geq \lambda_{2} \geq \cdots \geq \lambda_{2L} \geq 0$ if at least one connected path exists for any two nodes in $S$, guaranteed by above settings of the graph. The corresponding eigenvectors are denoted as $\Phi_{1}, \Phi_{2}, \cdots, \Phi_{2L}$. The diffusion distance $D^{t}(i,j)$ can be rewritten as
\begin{equation}
    D^{t}(i,j) = \sum^{2L}_{r=1} \lambda^{2t}_{r}(\Phi_{r}(i) - \Phi_{r}(j))^2
\end{equation}

Identical to the arrangement of transition matrix $P$ as shown in Fig. \ref{fig:inconsistency_matrix}, $D^{t}$ also characterizes two types of diffusion distances for real and fake videos: (1) the diffusion distances in the same direction, i.e., $D^{t}_{hh}$ and $D^{t}_{vv}$, which would serve as the pseudo labels in the proposed DA Loss in a weakly supervised way later in Section \ref{sec: dal}. (2) the diffusion distances cross the direction, i.e., $D^{t}_{hv}$ and $D^{t}_{vh}$, which are incorporated in the cross attention module to obtain discriminative spatiotemporal representations.

\subsubsection{Directional Cross Attention (DiCA)}
In our implementation of DiCA with six cross-attention layers, the diffusion distances $D^{t}_{hv}$ and $D^{t}_{vh}$ are also taken into account and serve as biases to capture more crucial directional clues. Concretely, we first transform $D^{t}_{hv}$, $D^{t}_{vh}$ to diffusion similarity matrix via a negative exponential function. For one DiCA block, given the directional sequences $Z_{h}$ and $Z_{v}$, the learnable parameters $W^{h}_{q}, W^{v}_{q}$ are used for query projection, $W^{h}_{k}, W^{v}_{k}$ for key projection, and $W^{h}_{v}, W^{v}_{v}$ for value projection in horizontal and vertical directions. Then, with the Softmax function $\phi$, we have:
\begin{gather}
        Z^{'}_{h} =\mathrm{\phi} \left( \frac{W^{v}_{q}Z_{v} \cdot (W^{h}_{k}Z_{h})^{\top}}{\sqrt{E}}+ \mathrm{exp}(-\tau_{1} \cdot D_{vh}^{t}) \right )W^{h}_{v}Z_{h}, \\
        Z^{'}_{v} =\mathrm{\phi} \left( \frac{W^{h}_{q}Z_{h} \cdot (W^{v}_{k}Z_{v})^{\top}}{\sqrt{E}}+ \mathrm{exp}(-\tau_{1} \cdot D_{hv}^{t}) \right )W^{v}_{v}Z_{v},
\end{gather}
where $Z^{'}_{h}$ and $Z^{'}_{v}$ are the output of the cross-attention layer, $\tau_{1}$ is a learnable scale parameter. The proposed pipeline of fusing directional inconsistency clues takes advantage of both directional interaction and cross-diffusion patterns, with the expectation of capturing discriminative inconsistency features.

\subsection{Spatiotemporal Data Augmentation}

Data Augmentation (DA) has long been regarded as an effective way to improve the performance of representation learning. Considering the fact that advanced generated algorithms are emerging, which could generate visually convincing facial images with much fewer artifacts. Therefore the deepfake detection schemes trained on existing datasets could only exhibit inferior generalization performance.

Inspired by the works in \cite{li_face_2020, shiohara_detecting_2022, wang_altfreezing_2023}, we also take advantages of data augmentation strategy to encourage the detector to capture more substantial artifacts. Specifically, we augment video samples in two ways. (1) Spatial DA. To allow the DIP to mine more crucial spatial forgery artifacts, we diminish the effects of common spatial artifacts, e.g., high-frequency artifacts, color mismatches, and noise patterns, by augmenting samples with Gaussian blur, color saturation, Gaussian noise, etc. (2) Temporal DA. Similarly, we also mitigate the negative effects of unreliable temporal artifacts, i.e., frame dropping and repeating, by augmenting samples with them.

Furthermore, to prevent the optimization of our DIP from being trapped with hard samples, i.e., augmented samples, during the early stage of model training, we adopt a student-teacher optimization framework \cite{NIPS2017_ema} as illustrated in Fig. \ref{fig:optim_overview}. Specifically, after conducting spatiotemporal DA, we denote the processed inputs as $V_{anc}$, which is not augmented, $V_{pos},$ and $V_{neg}$, where $V_{anc}$ and $V_{pos}$ are sampled from the same real video, and $V_{neg}$ is sampled from the corresponding forged video. Note that the backbone, illustrated in Fig. \ref{fig:overview}, is utilized for both student model $M_{s}$ and teacher model $M_{t}$ in our implementation. And $V_{anc}$ is fed forward to $M_{s}$, while $V_{pos}$ and $V_{neg}$ are encoded by $M_{t}$. Note that we apply different strategies to update the parameters $\theta_{s}$ of $M_{s}$ and the parameters $\theta_{t}$ of $M_{t}$, which are detailed in the following section.

\subsection{Loss Functions}
\begin{figure}
    \centering
    \includegraphics[width= \linewidth]{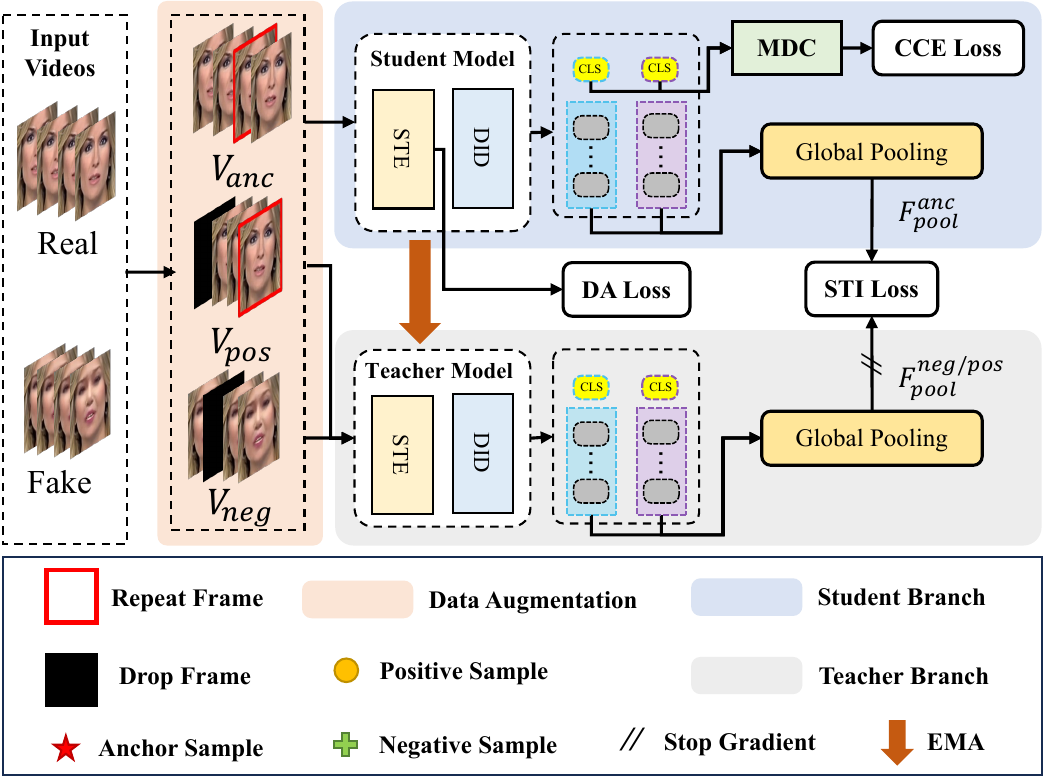}
    \caption{Overview of the proposed optimization framework.}
    \label{fig:optim_overview}
\end{figure}

As illustrated in Fig.\ref{fig:optim_overview}, the proposed DIP model follows a student-teacher framework, including spatiotemporal DA and three loss functions. In specific, the parameters of $M_{s}$ are updated with gradient backpropagation on loss functions, and the parameters of $M_{t}$ are weighted summations of the student and teacher models through the exponential moving average strategy \cite{NIPS2017_ema}, i.e.,

\begin{equation}
    \theta_{t} = \alpha \ast \theta_{t} + (1 - \alpha)\ast \theta_{s},
\end{equation}
where $\alpha$ is the momentum of updating the parameters $\theta_{t}$ and set as $0.99$.

\subsubsection{SpatioTemporal Invariant Loss (STI Loss)}
$L_{sti}$ is a loss function in alignment with the proposed spatiotemporal data augmentation strategy to mine general forgery features in fine-grained triplets by increasing intra-class compactness and inter-class separability. The loss function is defined as:
\begin{equation}
    L_{sti} = max(d_{sti} + cos(F_{pool}^{anc}, F_{pool}^{neg}) - cos(F_{pool}^{anc}, F_{pool}^{pos}), 0),
\label{eqution:triplet}
\end{equation}
where $d_{sti}$ is the degree of intra-class separability, set as 1.0, and $cos(\cdot)$ represents the cosine distance to measure the similarity of samples. $F_{pool}^{anc}, F_{pool}^{neg}$ and $F_{pool}^{pos}$ are globally pooled and directionally concatenated features for the positive, anchor, and negative samples. 

As shown in Eq. (\ref{eqution:triplet}), $L_{sti}$ constrains the similarity between the anchor/positive pair is greater than the one between the anchor/negative pair by at least  $d_{sti}$. Therefore, equipped with spatiotemporal DA and $L_{sti}$, the proposed model tends to learn more substantial spatiotemporal forgery clues for deepfake detection.

\subsubsection{Diffusion-Aware Loss (DA Loss)}
\label{sec: dal}
The prior knowledge of $Z_{d}$ $(d \in \{h,v\})$ is exploited to speed up the diffusion process in the IDM. Specifically, the DA loss $L_{da}$ is adopted to compute the normalized cosine similarity and diffusion distance among tokens in the same direction and encourage them to be close to each other. The DA loss can be written as:
\begin{equation}
    L_{da} = \sum_{d \in \{ h,v \}} ||norm(cos(Z_{d}, Z_{d}))- \mathrm{exp}(-\tau_{2} \cdot D_{dd}^{t}) ||^{2}_{2},
\end{equation}
where $norm(\cdot)$ is a function that normalizes the cosine similarities into $[0,1]$ and $\tau_{2}$ is a learnable scale parameter. Note that when the normalized cosine similarity of $Z_{d}$ increases, the exponentially scaled $D_{dd}^{t}$ is expected to be larger (i.e., smaller diffusion distance), and vice versa.

\subsubsection{Collaborative Cross-Entropy Loss (CCE Loss)}
As shown in Fig. \ref{fig:overview}, in addition to the video-based prediction, both the horizontal and vertical representations could also exhibit complementary directional forgery clues. Therefore, the CCE Loss is developed, which allows the proposed DIP model to extract discriminative inconsistency features globally and directionally. 
\begin{equation}
    L_{cce} = l(y, \hat{y}) + \sum_{d \in \{ h,v \}}\lambda_{d} \cdot l(y_{d}, \hat{y}),
\end{equation}
where $l(\cdot)$ represents the cross-entropy loss function, $y$ and $\hat{y}$ are the video-based prediction probability and ground truth, respectively. $y_{d}$ $(d \in \{h,v\})$ stands for directional prediction probability and $\lambda_{d}$ is the weight for each directional loss.

As a result, the overall loss function for the proposed DIP framework could be written as the sum of all the above loss functions, i.e.,

\begin{equation}
    L_{all} = L_{cce} + L_{sti} + L_{da}.
\end{equation}

\section{Experiments}
\subsection{Experimental Settings}
\subsubsection{Implementation Details}

For each video, 8 clips, each with 16 frames, are sampled for training and testing. The open-source Dlib algorithm is employed to detect and crop facial regions, which are then resized to $224 \times 224$. The proposed method is implemented using the PyTorch library on the four NVIDIA RTX 3090 platforms. For model training, the AdamW optimizer [42] with an initial learning rate of $4 \times 10^{-6}$ is utilized, and the batch size is set to 16. The setting of each transformer block in DIP follows the ViT model \cite{dosovitskiy_image_2021}. For the hyperparameter settings, the neighborhood number $k_{n}$ of one node and the diffusion step $t$ in IDM are set to $7, 20$, respectively. The optimal ratio of spatial and temporal attention layers in the STE block will be discussed in the following ablation experiments.

\subsubsection{Evaluation Metrics}
Following the evaluation protocols in \cite{haliassos_lips_2021, shiohara_detecting_2022}, the Area Under the Receiver Operating Characteristic Curve (AUC) and Detection Accuracy (ACC) at the video level are utilized as evaluation metrics. Specifically, for comparison with image-level methods, the video-level results of such methods are reported by averaging the detection scores of all frames.

\subsubsection{Dataset}
\textbf{FaceForensics++} (FF++) \cite{rossler_faceforensics_2019}: FF++ is a widely used benchmark in deepfake detection. It includes 1,000 real videos collected from YouTube and 4,000 fake videos generated by four types of face manipulation techniques: Deepfakes (DF) \cite{noauthor_deepfakes_2024}, Face2Face (F2F) \cite{thies_face2face_2016}, FaceSwap (FS) \cite{marek_faceswap_2020}, and NeuralTextures (NT) \cite{thies_deferred_2019}. In addition, videos in FF++ are compressed at three quality levels: raw, high-quality (C23), and low-quality (C40). To better simulate practical scenarios, the C23 and C40 versions are adopted to conduct our experiments.

\textbf{Celeb-DF-v2} (CDF) \cite{li_celeb-df_2020}: CDF is a large-scale public deepfake dataset that contains 590 real videos and $5,639$ high-quality face-swapped videos of celebrities. 

\textbf{WildDeepFake} (WDF) \cite{zi_wilddeepfake_2020}: WDF is a real-world dataset with $3,805$ real videos and $3,805$ fake videos containing diverse and complicated conditions of video collection and manipulation, which is more challenging for deepfake detection.

\textbf{Deepfake Detection Challenge} \cite{dolhansky_deepfake_2019, dolhansky_deepfake_2020} generates more challenging forged faces, including two versions: preview and complete datasets termed DFDC-P and DFDC,  which apply some specifically designed augmentations to the target videos to approximate the actual degradations.

\textbf{DeeperForensics-1.0} (DFR) \cite{jiang_deeperforensics-10_2020}: DFR applies more sophisticated forgery methods and well-controlled video sources to generate "natural" deepfake videos with various distortions.

\subsection{Intra-dataset Evaluation}
We first evaluate our proposed DIP with state-of-the-art methods under the intra-dataset setting, in which all models are trained and tested on FF++ with multiple qualities, i.e., C23 and C40. The results are shown in Table \ref{tab:intra-dataset}. Our DIP achieves a comparable and balanced detection performance compared with the SOTA methods across different video qualities.

As forgery artifacts are greatly weakened by strong video compression, it is more challenging to extract discriminative artifact features from highly compressed videos. Despite this,  our proposed DIP focuses on mining fine-grained spatiotemporal forgery artifacts and captures multidirectional forgery patterns, which is more robust in video compression, thus achieving comparable performance with various video qualities. For instance, compared with the recent image-based methods MRL \cite{yang_masked_2023}, our DIP improves the ACC from $93.82\%$ to $99.10\%$ for high-quality data while achieving satisfactory performance on challenging low-quality videos. The difference in detection performance on MRL could be caused by quality-sensitive artifacts, resulting in unbalanced performance on different video qualities. In contrast, both the proposed DIP and ISTVT \cite{zhao_istvt_2023} incorporate temporal forgery artifacts, which are more invariant to video compression and could capture discriminative clues even on low-quality videos.
\begin{table}[ht]
\centering
\caption{Intra-Dataset Evaluation on FF++. The best and second-best results are shown in bold and underlined text, respectively.}
\begin{tabular}{llcccc}
\toprule[1pt]
\multirow{2}{*}{Method} & \multirow{2}{*}{Venue} & \multicolumn{2}{l}{FF++ (C40)} & \multicolumn{2}{l}{FF++ (C23)} \\ \cline{3-6} 
 &  & ACC & AUC & ACC & AUC \\ \hline
Face X-ray \cite{li_face_2020}  & CVPR 2020 & - & 61.60 & - & 87.40 \\
F3Net \cite{qian_thinking_2020} & ECCV 2020 & 90.43 & 93.30 & 97.52 & 98.10 \\
FDFL \cite{li_frequency-aware_2021} & CVPR 2021 & 89.00 & 92.40 & 96.69 & 99.30 \\
DCL \cite{sun_dual_2022} & AAAI 2022 & - & - & 96.74 & 99.30 \\
UiA-ViT \cite{zhuang_uia-vit_2022} & ECCV 2022 & - & - & - & 99.33 \\
Lisiam \cite{wang_lisiam_2022} & TIFS 2022 & 91.29 & 94.65 & 97.57 & \textbf{99.52} \\
CDIN \cite{cdin_tcsvt} &TCSVT 2023 &- &\textbf{96.80} & - & 98.50 \\
MRL \cite{yang_masked_2023} & TIFS 2023 & 91.81 & 96.18 & 93.82 & 98.27 \\
ISTVT \cite{zhao_istvt_2023} & TIFS 2023 & \textbf{96.15} & - & \underline{99.00} & - \\
LDFNet \cite{lfdnet_tcsvt} & TSCVT 2024 & 92.32 & \underline{96.79} & 96.01 & 98.92 \\
DIP & Ours & \underline{92.33} & 95.16 & \textbf{99.10} & \underline{99.46} \\
\bottomrule[1pt]
\end{tabular}

\label{tab:intra-dataset}
\end{table}
\subsection{Generalizability Evaluation} 
In practical scenarios, the manipulation methods and data source of the suspicious face video to be detected are usually unknown, which requires the well-trained detector to capture highly generalized forgery patterns and exhibit good generalization ability. To evaluate the generalizability of the proposed method, we adopt two more practical settings: cross-dataset and cross-manipulation evaluations.

\subsubsection{Cross-dataset Evaluation} 

\begin{table*}[]
\centering
\caption{Generalizability Evaluation in terms of AUC (\%). All models are trained on FF++, and tested on the remaining five datasets, with video-level metrics. The best and second-best results are bolded and underlined, respectively.}
\begin{tabular}{lllcccccc}
\toprule[1pt]
Category & Method & Venue & CDF & DFDC-P & DFDC & DFR & WDF & Average \\ \hline
\multirow{7}{*}{Image-based} & Face X-ray \cite{li_face_2020} & CVPR 2020 & 79.50 & 74.20 & 65.50 & 86.80 & - & 76.50 \\
 & F3Net \cite{qian_thinking_2020} & ECCV 2020 & 68.69 & 67.45 & - & - & - & 68.07 \\
 & PCL \cite{zhao_learning_2021} & ICCV 2021 & $\textbf{90.03}$ & 74.37 & 67.52 & - & - & 77.31 \\
 & DCL \cite{sun_dual_2022} & AAAI 2022 & 88.24 & 77.57 & - & 94.42 & 76.87 & 84.28 \\
 & RECCE \cite{cao_end--end_2022} & CVPR 2022 & 69.25 & 66.90 & - & 93.28 & \underline{76.99} & 74.67 \\
 & SBI \cite{shiohara_detecting_2022} & CVPR 2022 & \underline{89.90} & \underline{86.15} & 72.42 & 77.70 & 70.27 & 79.78 \\
 & Lisiam \cite{wang_lisiam_2022} & TIFS 2022 & 78.21 & - & - & - & - & 78.21 \\ \hline
\multirow{7}{*}{Video-based} & FTCN \cite{zheng_exploring_2021} & ICCV 2021 & 86.90 & 71.00 & 74.00 & \textbf{98.80} & - & 82.70 \\
 & LipForensics \cite{haliassos_lips_2021} & CVPR 2021 & 82.40 & - & 73.50 & 97.60 & - & 84.50 \\
 & HCIL \cite{gu_hierarchical_2022} & ECCV 2022 & 79.00 & - & 69.21 & - & - & 74.11 \\
 & RATIL \cite{gu_region-aware_2022} & IJCAI 2022 & 76.50 & - & 69.06 & - & - & 72.78 \\
 & STDT \cite{zhang_deepfake_2022} & MM 2022 & 69.78 & - & 66.99 & - & - & 68.39 \\
 & CDIN \cite{cdin_tcsvt} & TCSVT 2023 & 89.10 & - & 78.40 & - & - & 83.75 \\
 & AltFreezing \cite{wang_altfreezing_2023} & CVPR 2023 & 89.50 & - & 71.25 & 99.30 & - & \underline{86.68} \\
 & AdapGRnet \cite{guo_exposing_2023} & TMM 2023 & 71.50 & - & - & - & - & 71.50 \\
 & ISTVT \cite{zhao_istvt_2023} & TIFS 2023 & 84.10 & - & \underline{74.20} & \underline{98.60} & - & 85.63 \\ \cline{2-9} 
\multicolumn{1}{l}{} & DIP & Ours & 88.36 & \textbf{87.98} & \textbf{79.90} & 98.02 & \textbf{81.12} & \textbf{87.08} \\ 
\bottomrule[1pt]
\end{tabular}
\label{tab:generalization}
\end{table*}

Table \ref{tab:generalization} shows the generalization results for cross-dataset evaluation in terms of AUC, in which all models are trained in FF++ (C23) for a fair comparison. It can be observed that our DIP exhibits superior performance in most cases, including the average result compared with the prior methods. in specific, our method achieves significant performance improvements of 4.13\% and  5.70\% in AUC on WDF and DFDC, respectively, compared with state-of-the-art methods, such as DCL \cite{sun_dual_2022}, SBI \cite{shiohara_detecting_2022}, RECCE \cite{cao_end--end_2022}, and ISTVT \cite{zhao_istvt_2023}. Previous image-based methods commonly suffer from dramatic performance drops when evaluated on DFDC, WDF, and DFR, in which videos are distorted with complicated perturbations and thus more challenging to detectors, while the proposed DIP takes full advantage of temporal artifact mining and achieves the best detection performance, indicating temporal artifacts are significant and non-negligible clues against highly distorted videos in the cross-dataset evaluation.

Following the general spatiotemporal learning paradigm, the FTCN \cite{zheng_exploring_2021} and ISTVT \cite{zhao_istvt_2023} explore temporal forgery artifacts for deepfake detection and achieve better performance on CDE and DFR compared with the image-level methods. However, they fail to consider the importance of multi-directional forgery pattern modeling, which results in poor generalization on challenging cases, e.g., WDF and DFDC. It is benefiting from multidirectional forgery pattern modeling and temporal data augmentation that our DIP could achieve better cross-dataset detection performance across all datasets.

\subsubsection{Cross-manipulation Evaluation}
Following previous works \cite{haliassos_lips_2021, yu_augmented_2023}, we then proceed to evaluate the cross-manipulation performance of the proposed DIP on FF++. Specifically, all models are trained on three manipulation methods and tested on the remaining one on FF++. It can be observed in Table \ref{tab:cross-manipulation}, when tested on FS, our proposed DIP outperforms other counterparts with a near $3 \%$ AUC improvement. Compared to the state-of-the-art DIL \cite{gu_delving_2022}, our DIP consistently achieves better average cross-manipulation performance. The generalization performance of DIL benefits from temporal and local forgery artifact mining, especially on NT. However, it gains unsatisfactory performance when applied to the other three cross-manipulation evaluations since it fails to capture the global forgery artifacts. In contrast, our DIP could effectively mine both global and local forgery artifacts and capture more discriminative forgery patterns, resulting in better generalizability in various cross-manipulation scenarios.

\begin{table}[]
\centering
\caption{Cross-manipulation Evaluation on four forgery methods in terms of AUC (\%). All mentioned models are trained on three manipulation methods and tested on the remaining one on FF++. The best and second-best results are bolded and underlined, respectively.}
\begin{tabular}{llccccc}
\toprule[1pt]
\multirow{2}{*}{Method} & \multirow{2}{*}{Venue} & \multicolumn{5}{c}{Train on remaining three} \\ \cline{3-7} 
 &  & DF & FS & F2F & NT & \textbf{Avg} \\ \hline
Xcep. \cite{chollet_xception_2017} & ICCV 2017 & 93.9 & 51.2 & 86.8 & 79.7 & 77.9 \\
Lips \cite{haliassos_lips_2021} & CVPR 2021 & 93.0 & 56.7 & \underline{98.8} & \textbf{98.3} & 86.7 \\
FTCN \cite{zheng_exploring_2021} & CVPR 2021 & 96.7 & \underline{96.0} & 96.1 & \underline{96.2} & \underline{96.3} \\
DIAnet \cite{hu_dynamic_2021} & IJCAI 2021 & 87.9 & 86.1 & 88.2 & 86.7 & 87.2 \\
DIL \cite{gu_delving_2022} & AAAI 2022 & 94.4 & 94.8 & 94.8 & 94.9 & 94.7 \\ \hline
DIP & Ours & \underline{98.2} & \textbf{99.0} & \textbf{99.1} & 92.4 & \textbf{97.2} \\ 
\bottomrule[1pt]
\end{tabular}
\label{tab:cross-manipulation}
\end{table}

\subsection{Robustness Evaluation}
\begin{figure*}[h]
    \centering
    \includegraphics[width= 0.8\textwidth]{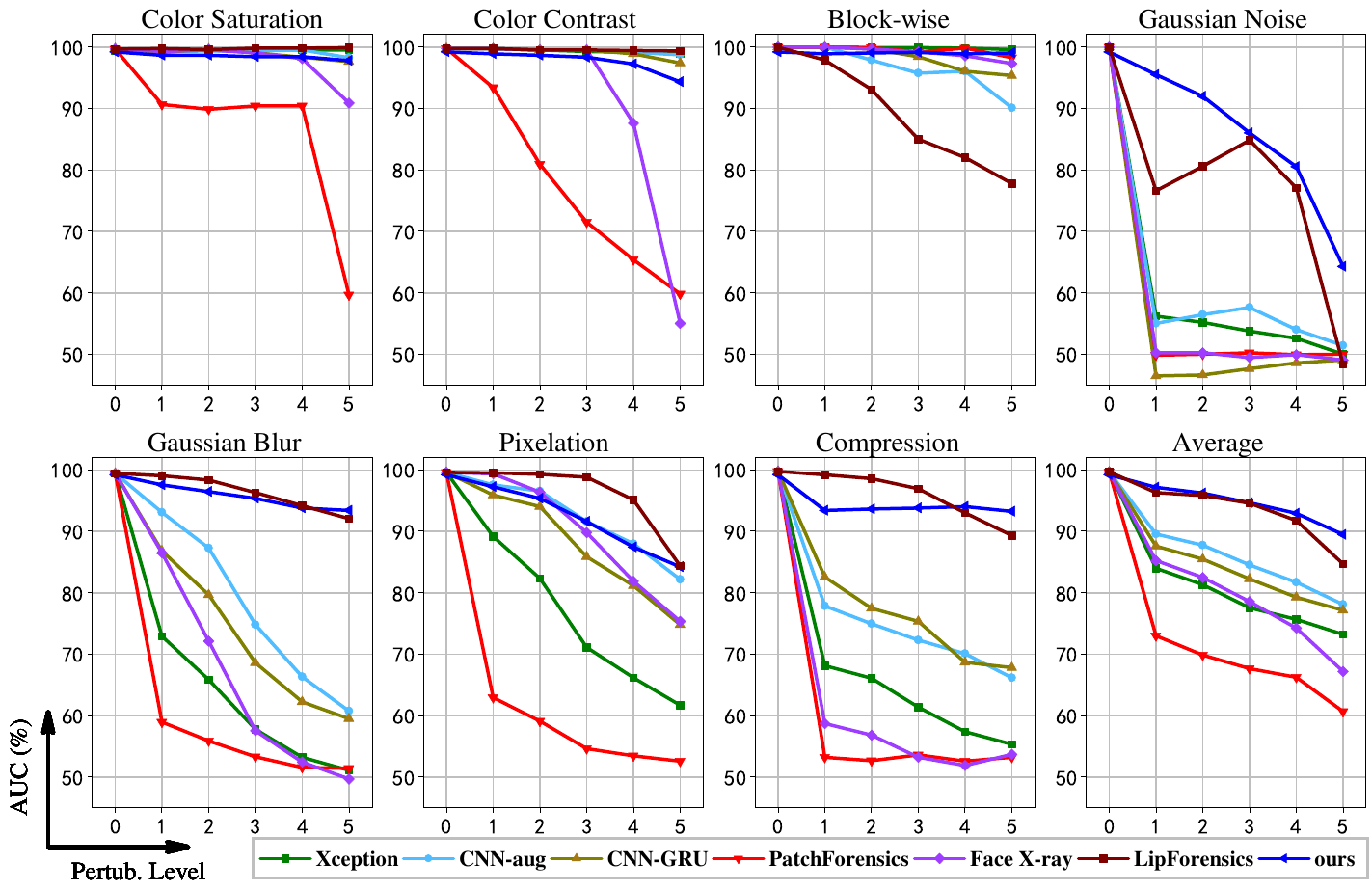}
    \caption{Robustness Evaluation within unseen distortions of various levels in terms of AUC (\%). Several models trained on \emph{clean} FF++ are evaluated under several distortions with different levels. "Average" denotes the average performance against the same-level distortions.}
    \label{fig:robustness}
\end{figure*}

During collection and transmission, the video is usually distorted with unknown intensities, which would potentially weaken or destroy essential forgery artifacts. To evaluate the robustness of the proposed method, following the previous work \cite{haliassos_lips_2021}, we perform robustness evaluations on our DIP and other compared methods under multiple distortions with different intensities, including color saturation modification, color contrast modification, blockwise noise, Gaussian noise, Gaussian blur, pixelation, and video compression. 

As shown in Fig. \ref{fig:robustness}, our proposed DIP is more robust than the other involved methods under various distortion scenarios. Since the spatial forgery artifacts are heavily corrupted under spatial distortions, including Gaussian noise, pixelation, color saturation, and contrast, while the interframe relationships, i.e. temporal artifacts, are relatively well preserved,   the video-based methods,  i.e., LipForensics \cite{haliassos_lips_2021} and the proposed DIP, are significantly more robust than the image-based methods, e.g., Xception \cite{rossler_faceforensics_2019}, Face X-ray \cite{li_face_2020}, and PatchForensics \cite{chai_what_2020}, which rely heavily on spatial forgery artifacts.

Video compression corrupts temporal forgery artifacts to a great extent. In this scenario, our proposed DIP exhibits better robustness compared with LipForensics. The reason is that LipForensics fails to model the global temporal patterns with the constraint of local forgery pattern modeling. In contrast, our DIP considers both local and global forgery artifacts and implements multidirectional forgery pattern modeling, thereby achieving more robust detection performance. 

\subsection{Ablation Studies}
We conduct extensive ablation experiments to analyze the impacts of various components in the proposed DIP. All models and their variants are trained on FF++ and evaluated on intra-dataset and cross-dataset detection performance in terms of AUC (\%).

\subsubsection{Study on the architectures of STE unit}
To determine the optimal setting in STE, i.e., the ratio to the number of spatial attention (SA) and temporal attention (TA) modules in one unit, we also conduct experiments under different settings, and the results are shown in Table \ref{table:STE unit}. The DIP framework is not expected to gain detection performance improvements by stacking more temporal attention modules, and adequate spatial forgery pattern learning with increasing spatial attention modules could promote spatiotemporal inconsistency pattern learning. Moreover,  the setting of three SA modules can better balance spatial and temporal inconsistency pattern learning, which is chosen as the default setting of DIP.

\subsubsection{Study on the effects of the STE}
We further investigate three paradigms of deepfake spatiotemporal learning, i.e., spatial-then-temporal learning (FTCN \cite{zheng_exploring_2021}), one-spatial-one-temporal learning (ViViT \cite{arnab_vivit_2021}), and more-spatial-one-temporal learning (Ours). The FTCN extracts frame-level forgery representations in the first stage and then focuses on capturing temporal forgery artifacts, which fails to mine the importance of local forgery artifacts and results in unsatisfactory detection performance. To better capture local spatiotemporal forgery artifacts, ViViT fuses spatiotemporal forgery learning by factorizing self-attention layers into spatial and temporal ones in equal numbers. Furthermore, the proposed DIP designs asymmetric spatiotemporal attention. It can be observed in Table \ref{table:STE} that the asymmetric spatiotemporal attention design in the proposed DIP could better learn spatiotemporal forgery representations and achieve significant performance improvements.

\subsubsection{Study on different settings of $k_{n}$ and $t_{d}$}
At the inconsistency diffusion module, $k_{n}$ in Eqn. (\ref{eqution: neighbor}) controls the scale of the node's neighborhood in the graph $G$, while $t$ in Eqn. (\ref{eqution: diffu}) constrains diffusion steps. Table \ref{tab:hyperparameters_ablation} presents the performance under various combinations of $k_{n}$ and $t$. Initially, we fix $k_{n}$ to 3 or 7 to examine the impacts of different $t$ values. We observe that when $t$ is set as 20, the overall performance is more balanced. With respect to $k_{n}$,  when fewer connected neighboring nodes are involved in the inconsistency transition graph, insufficient forgery relationships prevent this module from modeling more discriminative forgery diffusion patterns, thereby gaining unsatisfactory detection performance. However, a much larger $k_{n}$ would result in the loss of localities of inconsistency relationships, leading to lower cross-dataset performance.

\subsubsection{Study on the effects of DID}
We then proceed to study the impacts of DiCA, IDM, and DA Loss. The results are shown in Table \ref{table:Biid}. When integrated with only DiCA or IDM, the generalizability performance is improved, which indicates that the directional interaction (DiCA) and diffusion (IDM) are vital for inconsistency modeling. Moreover, the IDM achieves better performance on the CDF, which indicates that the CDF exhibits more discriminative inconsistency diffusion patterns, rather than directional interaction patterns compared with DFDC-P. When equipped with all the three components, both the intra-dataset and cross-dataset performance are further boosted, indicating that our proposed DIP could mine more various forgery patterns and generalize well on cross-dataset evaluations.

\subsubsection{Study on the effects of Spatiotemporal DA}
Spatiotemporal DA methods designed with forgery-related prior knowledge are essential to prevent the detector from overfitting nonintrinsic patterns. To measure the impacts of all the involved DA methods, we train several variants of the proposed DIP, including models without any one specific type of augmentation and one model without any augmentations. Additionally, several postprocessed FF++ datasets are involved in experiments to simulate different video perturbations, i.e., FFhf (FF++ with high-frequency filtering), FFnoise (FF++ with noise addition), FFcolor (FF++ with color distortion), and FFta (FF++ with random frame dropping and repeating).

Table \ref{tab:ablation-aug} summarizes the detection performance of all the settings in the FF++ dataset. It can be observed that the trained detector without data augmentations is more vulnerable to various distortions, especially high-frequency filtering and temporal distortions. Moreover, benefiting from the proposed data augmentation methods, our DIP further improves the generalization performance against various natural distortions, which indicates the effectiveness of the proposed spatiotemporal DA methods.

\begin{table}[h]
\centering
\caption{Study on the architectures of a unit block in STE. Note that four forgery methods in FF++(LQ) and CDF are leveraged to evaluate different settings of SA: TA ratios.}
\begin{tabular}{ccccccc}
\toprule[1pt]
SA & TA & DF & F2F & FS & NT & CDF\\ \hline
1 & 3 & 96.1 & 89.6 & 93.2 & 89.4 & 79.7 \\
1 & 1 & 98.2 & 92.9 & 95.4 & 93.2 & 83.1 \\
2 & 1 & 97.6 & 92.4 & 96.5 & 89.8 & 85.6 \\
3 & 1 & 98.9 & 95.7 & 96.3 & 95.5 & 88.3 \\
4 & 1 & 99.3  & 94.5 & 94.2 & 92.4 &  87.5\\
6 & 1 & 98.5 & 93.8 & 93.7 & 92.5 & 86.4 \\
\bottomrule[1pt]
\end{tabular}
\label{table:STE unit}
\end{table}

\begin{table}[h]
\centering
\caption{Study on the effects of STE. We compare performance with three paradigms of existing forgery spatiotemporal learning on intra-dataset and cross-dataset evaluations. Note that the best results are bolded.}
\begin{tabular}{llcccc}
\toprule[1pt]
Methods & Venue & FF++ & CDF & DFDC-P & Average \\ \hline
FTCN \cite{zheng_exploring_2021} & CVPR 2021 & \textbf{99.7} & 86.9 & 74.0 & 86.9 \\
ViViT \cite{arnab_vivit_2021} & ICCV 2021 & 97.9 & 80.5 & 74.2 & 84.2 \\
DIP & Ours & 99.4 & \textbf{88.3} & \textbf{88.0} & \textbf{92.9} \\
\bottomrule[1pt]
\end{tabular}
\label{table:STE}
\end{table}

\begin{table}[h]
\centering
\caption{Study on different settings of $k_{n}$ and $t$. Cross Avg. indicates average generalization performance at the AUC metric under cross-dataset scenarios (CDF, DFDC-P, and WDF).}
\begin{tabular}{ccccccc}
\toprule[1pt]
\multicolumn{2}{c}{Hyper-parameters} & \multirow{2}{*}{FF++} & \multirow{2}{*}{CDF} & \multirow{2}{*}{DFDC-P} & \multirow{2}{*}{WDF} & \multirow{2}{*}{Cross Avg.} \\ \cline{1-2}
$k_{n}$ & $t$ &  &  &  &  &  \\ \hline
3 & 10 & 99.33 & 84.11 & 86.25 & 79.81 & 83.39 \\
3 & 20 & 99.37 & 85.07 & 86.10 & 80.01 & 83.73 \\
7 & 20 & 99.46 & \textbf{88.36} & 87.98 & \textbf{81.12} & \textbf{85.82} \\
7 & 25 & 99.51 & 87.98 & 87.52 & 81.01 & 85.50 \\
7 & 40 & 99.42 & 88.45 & \textbf{88.23} & 80.24 & 85.64 \\
9 & 25 & 99.64 & 88.12 & 86.58 & 79.89 & 84.86 \\
15 & 25 & \textbf{99.82} & 87.40 & 86.38 & 80.76 & 84.85 \\ 
\bottomrule[1pt]
\end{tabular}
\label{tab:hyperparameters_ablation}
\end{table}

\begin{table}[h]
\caption{Study on the effects of DID. We decompose DID into several elements, i.e., DiCA, IDM, and DAL, and evaluate their contributions in intra-dataset (FF++) and cross-dataset (CDF and DFDC-P) settings.}
\centering
\begin{tabular}{cccccc}
\toprule[1pt]
DiCA & IDM & DA Loss & FF++ & CDF & DFDC-P \\ \hline
$\times$ & $\times$ & $\times$ & 99.2 & 84.3& 79.8 \\
$\checkmark$ & $\times$ & $\times$ & 99.6 & 85.1 & 84.6 \\
$\times$ & $\checkmark$ & $\times$ & 99.4 & 86.4 & 80.9 \\
$\times$ & $\checkmark$ & $\checkmark$ & 99.4 & 87.0 & 81.8 \\
$\checkmark$ & $\checkmark$ & $\times$ & 99.3 & 87.5& 85.2 \\
$\checkmark$ & $\checkmark$ & $\checkmark$ & 99.6 & 88.3 & 88.6 \\ 
\bottomrule[1pt]
\end{tabular}
\label{table:Biid}
\end{table}

\begin{table}[h]
\centering
\caption{Study on the effect of Spatiotemporal DA. Variants of the proposed DIP with different settings are evaluated in terms of AUC (\%): 'HFreq Aug.' (High-Frequency Augmentation), 'Noise Aug.' (Noise Augmentation), 'Color Aug.' (Color distorting Augmentation), and 'Tempo Aug.' (Temporal Augmentation). Note that the best results are bolded.}
\begin{tabular}{lccccc}
\toprule[1pt]
Variants & FF++ & FFhf & FFnoise & FFcolor & FFta \\ \hline
w/o Aug. & 99.23 & 92.71 & 95.60 & 96.26 & 91.89 \\
w/o HFreq Aug. & 99.16 & 93.66 & 97.13 & 98.01 & 97.57 \\
w/o Noise Aug. & 99.37 & \textbf{97.02} & 97.12 & 98.51 & 98.06 \\
w/o Color Aug. & 99.09 & 95.69 & 97.03 & 97.89 & 98.18 \\
w/o Tempo Aug. & 99.42 & 96.10 & 96.93 & 97.18 & 93.57 \\
Original & \textbf{99.46} & 96.60 & \textbf{98.91} & \textbf{98.53} & \textbf{98.62} \\
\bottomrule[1pt]
\end{tabular}
\label{tab:ablation-aug}
\end{table}

\subsection{Visualization}
In this subsection, we visualize directional saliency maps of input videos to better show the effectiveness of the directional patterns in our proposed DIP. Moreover, feature distributions and forgery patterns generated from DiCA and IDM are also leveraged to illustrate the superior forgery discrimination ability of our method.

\begin{figure}[]
    \centering
    \includegraphics[width=0.9\linewidth]{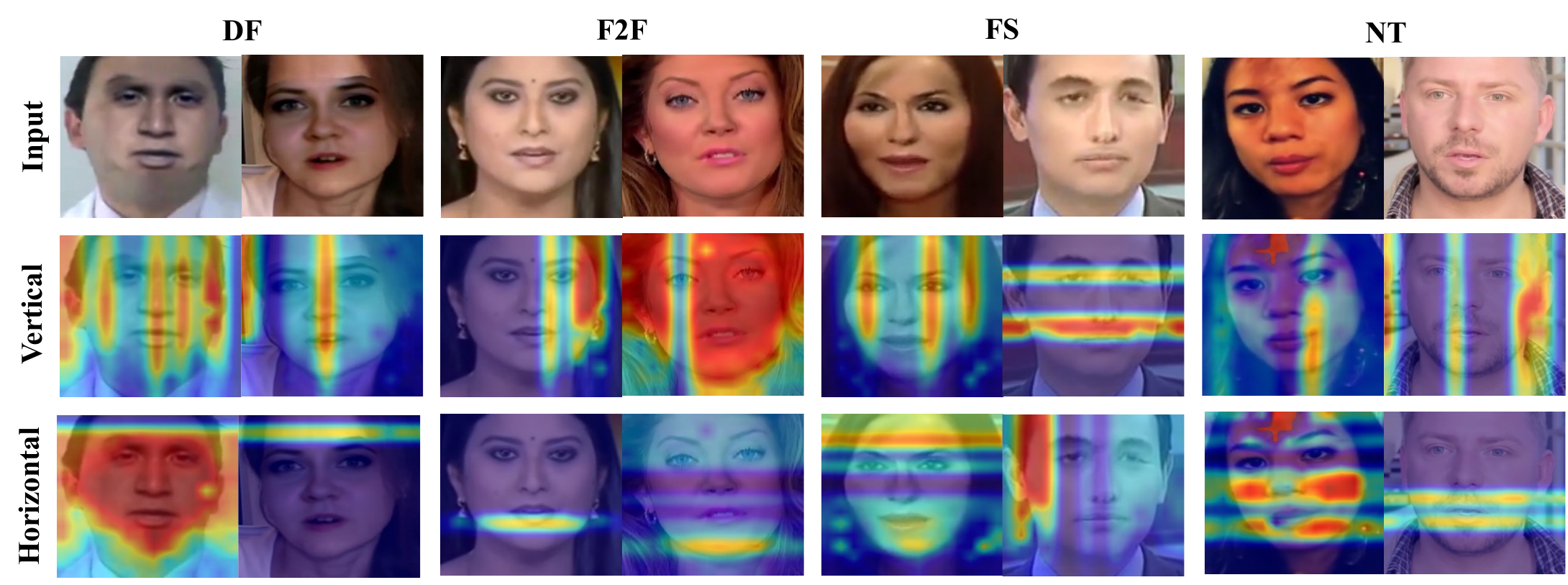}
    \caption{Visualization of horizontal and vertical forgery inconsistencies in the four facial manipulation algorithms. `Vertical' and `Horizontal' indicate the heatmaps in the vertical and horizontal directions, respectively.}
    \label{fig:directionAttn_visual}
\end{figure}

\begin{figure}[]
    \centering
    \includegraphics[width=\linewidth]{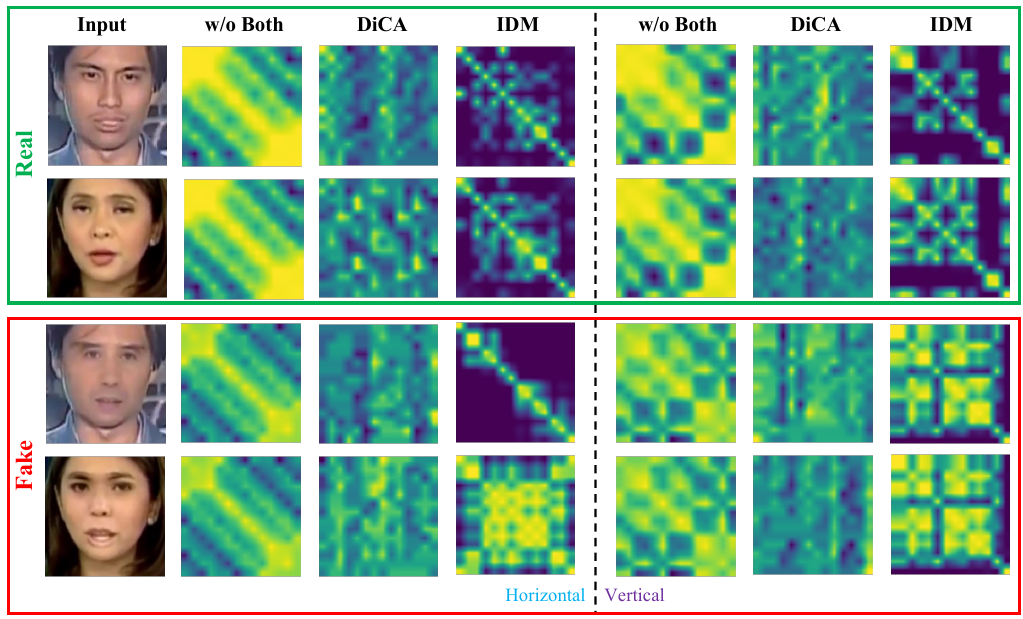}
    \caption{Visualization of inconsistency patterns modeled by DiCA and IDM. 'w/o Both' represents the inconsistency clues without pattern modeling by DiCA and IDM.}
    \label{fig:visualization_patterns}
\end{figure}

\begin{figure}[]
    \centering
    \includegraphics[width=\linewidth]{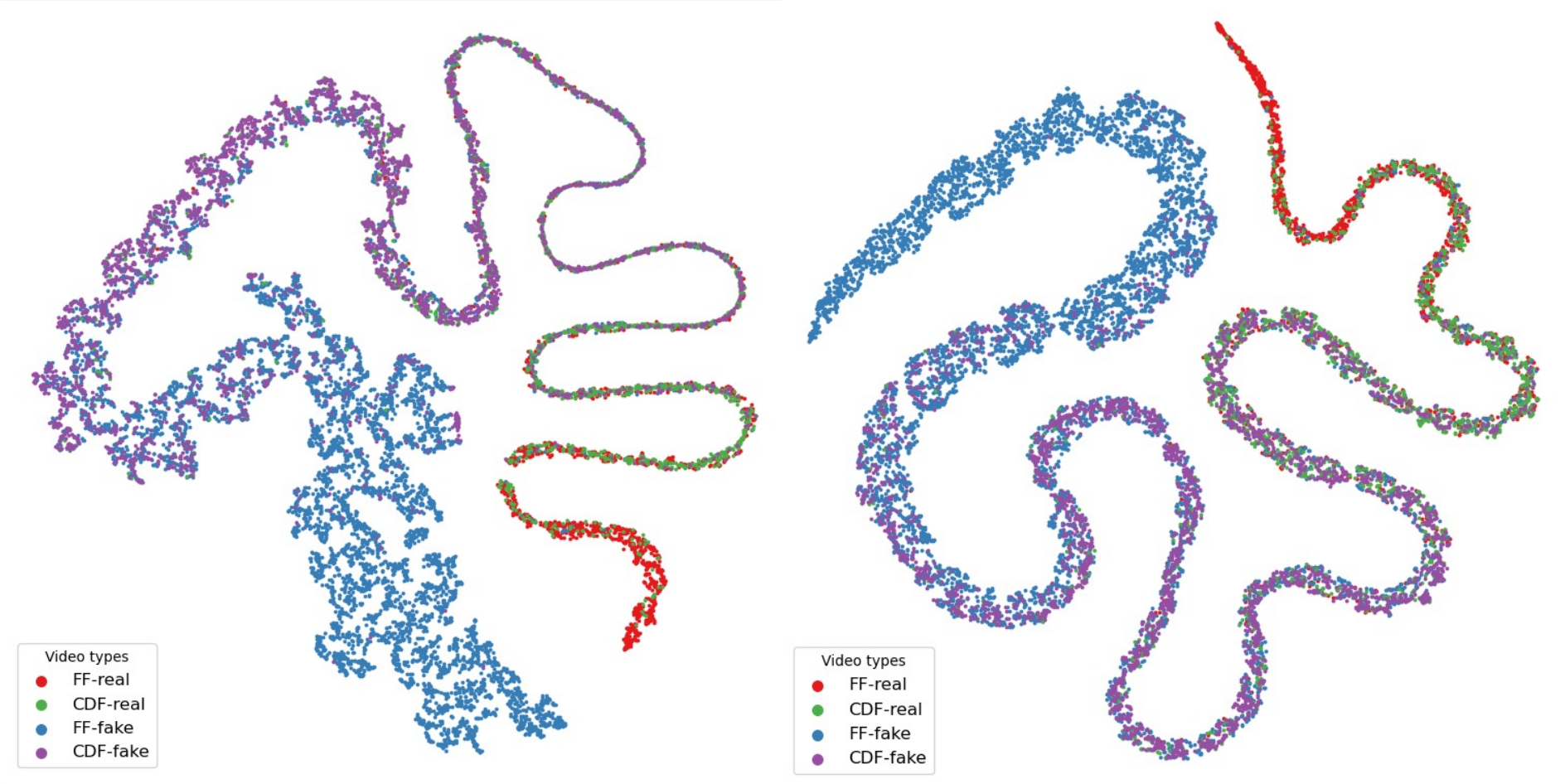}
    \caption{Visualization of feature distributions using t-SNE projection. \textbf{Left}: our DIP. \textbf{Right}: DIP without DiCA and IDM. Note that all models are trained on FF++.}
    \label{fig:visualization_tsne}
\end{figure}

\begin{figure}[]
    \centering
    \includegraphics[width=0.8\linewidth]{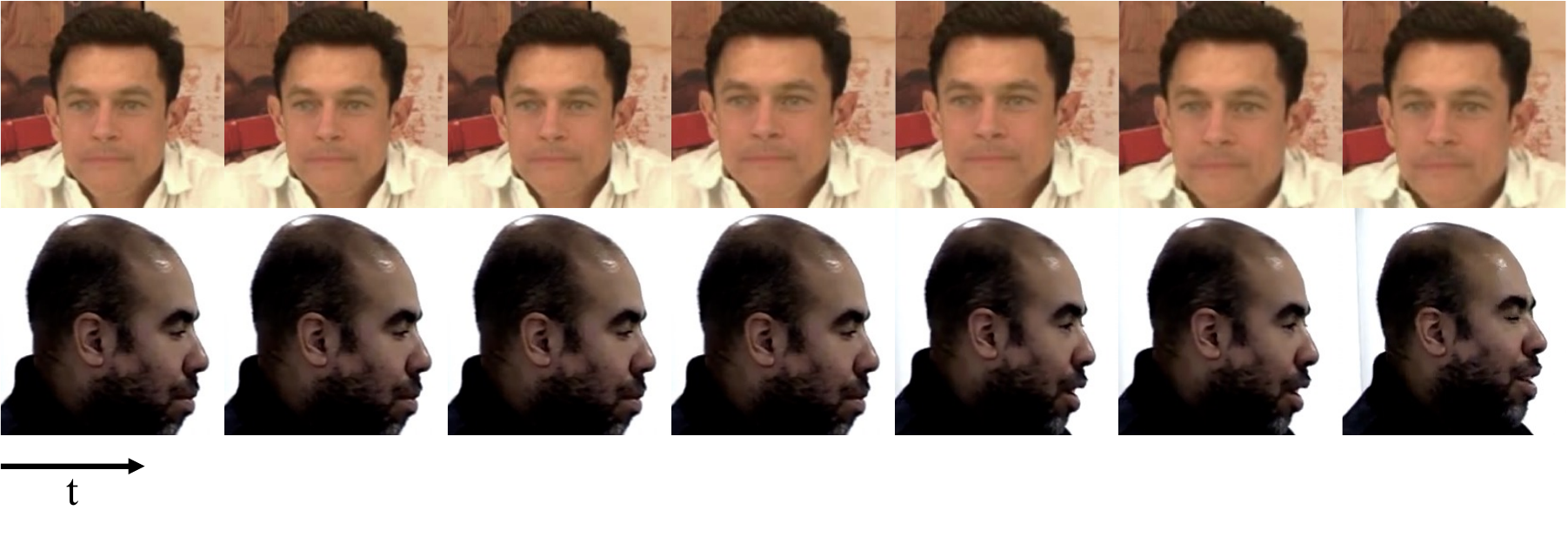}
    \caption{Examples of detection failure with the proposed method. \textbf{Top}: the face-swapping video (classified as real). \textbf{Bottom}: the real talking face video (classified as fake).}
    \label{fig:failure_case}
\end{figure}

\subsubsection{Directional Saliency Map Visualization}
We apply Grad-CAM \cite{selvaraju_grad-cam_2017} on horizontal and vertical directions of manipulated face clips to visualize where the proposed DIP pays attention to the forged faces in different directions. As shown in Fig.\ref{fig:directionAttn_visual}, spatiotemporal inconsistencies caused by deepfake methods have direction-aware characteristics, and our model is capable of discriminating directional forgery inconsistencies. In addition, the proposed DIP can learn the different characteristics of each deepfake method. In NeuralTextures (NT), which reenacts regions related to the mouth, the corresponding saliency maps show that forgery inconsistencies are located around the mouth region. In FaceSwap (FS), our model focuses on blending facial boundaries from different directions. These results provide a reasonable and straightforward explanation for the effectiveness of directional modeling in the DIP.

\subsubsection{Inconsistency Pattern Visualization}

Inconsistency patterns modeled by our DIP express regional embedding relationships from two different perspectives, i.e., BiCA and IDM. Here, we directly visualize the mentioned patterns to illustrate the effectiveness of the designed modules. As shown in Fig. \ref{fig:visualization_patterns}, compared with 'w/o Both', both the DiCA and IDM extract distinct patterns along both the horizontal and vertical directions. The IDM extracts more discriminative patterns than DiCA does. Specifically, patterns modeled by IDM demonstrate that inconsistency diffusion caused by deepfake algorithms is distributed evenly across regions, whereas diffusion caused by motion is more regionally independent, as DiCA models inconsistency patterns via a cross attention mechanism, i.e., directional interaction. Compared with forged videos, real videos have a more discrete pattern distribution. Therefore, the differences in inconsistency patterns between real and forged videos demonstrate the effectiveness of our DIP and reveal the forgery inconsistency patterns caused by existing deepfake techniques.

\subsubsection{T-SNE Feature Visualization}
We employ the t-SNE projection \cite{van2008visualizing} to analyze the proposed DIP and its variant in intra-dataset and cross-dataset scenarios. Fig. \ref{fig:visualization_tsne} shows different distributions of feature embeddings on both FF++ and CDF. For the intra-dataset evaluation (FF++), both DIP and DIP-variant could effectively distinguish manipulated samples from pristine ones and exhibit clear decision boundaries in the feature distribution. However, without the proposed DiCA and IDM, the DIP-variant fails to effectively discriminate between pristine faces and manipulated faces in the cross-dataset scenario (CDF). Particularly, some real faces in the CDF closely cluster with the manipulated faces from FF++ in the feature distribution. Moreover, there exist certain regions where features of all faces are clustered together, which negatively affects the generalization performance. Conversely, our DIP clusters pristine faces with a tighter state and maintain a high detection accuracy in the cross-dataset settings. Benefiting from explicit forgery pattern modeling implemented by DiCA and IDM, more universal forgery artifacts are captured and taken into final predictions, thereby verifying the effectiveness of both involved modules.

\subsubsection{Limitation}
While the proposed DIP achieves promising detection performance, especially on cross-dataset evaluation, there still exist a few failure instances with the proposed method. Fig. \ref{fig:failure_case} illustrates two video samples, where the face-swapping video is mispredicted as real (top), and the real talking face video (bottom) is misclassified as fake. This is because either the inconspicuous facial movements (top) or the extreme shooting angle (bottom) prevent the videos from exhibiting sufficient facial motion information. In other words, the proposed DIP could hardly capture their discriminative forgery clues for deepfake detection. And this also motivates us to delve into more general and robust forgery representation for deepfake detection, which remains as the topic of our future research efforts.

\section{Conclusion}
In this paper, we propose a novel spatiotemporal inconsistency learning method, i.e., \textbf{D}iffusion Learning of \textbf{I}nconsistency \textbf{P}attern (\textbf{DIP}), to exploit the directional spatiotemporal inconsistency for deepfake video detection by modeling inconsistency patterns from two perspectives, i.e., directional interaction and diffusion. These patterns are leveraged to learn more discriminative inconsistency clues jointly. In addition, the STI Loss is developed to facilitate the learning of invariant forgery information with well-devised spatiotemporal data augmentations, resulting in a generalized representation for deepfake detection. The experimental results show that our method outperforms other state-of-the-art methods on four commonly used benchmarks. Ablation studies also demonstrate the effectiveness of the involved designs.

\bibliographystyle{IEEEtran}

\begin{thebibliography}{10}
\providecommand{\url}[1]{#1}
\csname url@samestyle\endcsname
\providecommand{\newblock}{\relax}
\providecommand{\bibinfo}[2]{#2}
\providecommand{\BIBentrySTDinterwordspacing}{\spaceskip=0pt\relax}
\providecommand{\BIBentryALTinterwordstretchfactor}{4}
\providecommand{\BIBentryALTinterwordspacing}{\spaceskip=\fontdimen2\font plus
\BIBentryALTinterwordstretchfactor\fontdimen3\font minus \fontdimen4\font\relax}
\providecommand{\BIBforeignlanguage}[2]{{%
\expandafter\ifx\csname l@#1\endcsname\relax
\typeout{** WARNING: IEEEtran.bst: No hyphenation pattern has been}%
\typeout{** loaded for the language `#1'. Using the pattern for}%
\typeout{** the default language instead.}%
\else
\language=\csname l@#1\endcsname
\fi
#2}}
\providecommand{\BIBdecl}{\relax}
\BIBdecl

\bibitem{noauthor_deepfakes_2024}
\BIBentryALTinterwordspacing
``Deepfakes faceswap,'' Jan. 2024. [Online]. Available: \url{https://github.com/deepfakes/faceswap}
\BIBentrySTDinterwordspacing

\bibitem{thies_face2face_2016}
J.~Thies, M.~Zollhofer, M.~Stamminger, C.~Theobalt, and M.~Niessner, ``{Face2Face}: {Real}-{Time} {Face} {Capture} and {Reenactment} of {RGB} {Videos},'' in \emph{Proc. {IEEE} {Conf.} {Comput.} {Vis.} {Pattern} {Recognit.}}, Jun. 2016.

\bibitem{marek_faceswap_2020}
\BIBentryALTinterwordspacing
K.~Marek, ``{FaceSwap},'' 2020. [Online]. Available: \url{https://github.com/MarekKowalski/FaceSwap}
\BIBentrySTDinterwordspacing

\bibitem{wang_high-fidelity_2022}
T.~Wang, Y.~Zhang, Y.~Fan, J.~Wang, and Q.~Chen, ``High-{Fidelity} {GAN} {Inversion} for {Image} {Attribute} {Editing},'' in \emph{2022 {IEEE}/{CVF} {Conf.} {Comput.} {Vis.} {Pattern} {Recognit.}}, 2022, pp. 11\,369--11\,378.

\bibitem{thies_deferred_2019}
J.~Thies, M.~Zollhöfer, and M.~Nießner, ``Deferred {Neural} {Rendering}: {Image} {Synthesis} {Using} {Neural} {Textures},'' \emph{ACM Trans. Graph.}, vol.~38, no.~4, Jul. 2019.

\bibitem{wedel_improved_2009}
A.~Wedel, T.~Pock, C.~Zach, H.~Bischof, and D.~Cremers, ``An {Improved} {Algorithm} for {TV}-{L1} {Optical} {Flow},'' in \emph{Statistical and {Geo.} {Approach.} to {Vis.} {Moti.} {Analysis}}, 2009, pp. 23--45.

\bibitem{afchar_mesonet_2018}
D.~Afchar, V.~Nozick, J.~Yamagishi, and I.~Echizen, ``{MesoNet}: a {Compact} {Facial} {Video} {Forgery} {Detection} {Network},'' in \emph{{IEEE} {Int.} {Workshop} {Inf.} {Forensics} {Secur.}}, 2018, pp. 1--7.

\bibitem{rossler_faceforensics_2019}
A.~Rossler, D.~Cozzolino, L.~Verdoliva, C.~Riess, J.~Thies, and M.~Niessner, ``{FaceForensics}++: {Learning} to {Detect} {Manipulated} {Facial} {Images},'' in \emph{Proc. {IEEE}/{CVF} {Int.} {Conf.} {Comput.} {Vis.}}, 2019, pp. 1--11.

\bibitem{amerini_deepfake_2019}
I.~Amerini, L.~Galteri, R.~Caldelli, and A.~Del~Bimbo, ``Deepfake {Video} {Detection} through {Optical} {Flow} {Based} {CNN},'' in \emph{Proc. {IEEE}/{CVF} {Int.} {Conf.} {Comput.} {Vis.} {Workshops}}, 2019, pp. 0--0.

\bibitem{gu_spatiotemporal_2021}
Z.~Gu and et~al., ``Spatiotemporal {Inconsistency} {Learning} for {DeepFake} {Video} {Detection},'' in \emph{Proc. 29th {ACM} {Int.} {Conf.} {Multimedia}}, 2021, pp. 3473--3481.

\bibitem{zhang_detecting_2021}
D.~Zhang, C.~Li, F.~Lin, D.~Zeng, and S.~Ge, ``Detecting {Deepfake} {Videos} with {Temporal} {Dropout} {3DCNN}.'' in \emph{Proc. 30th {Int} {Joint} {Conf.} {Artif.} {Intell.}}, 2021, pp. 1288--1294.

\bibitem{gu_region-aware_2022}
Z.~Gu, T.~Yao, C.~Yang, R.~Yi, S.~Ding, and L.~Ma, ``Region-aware temporal inconsistency learning for deepfake video detection,'' in \emph{Proc. 31th {Int} {Joint} {Conf} on {Artif} {Intell}}, vol.~1, 2022, pp. 920--926.

\bibitem{guan_delving_2022}
J.~Guan and et~al., ``Delving into {Sequential} {Patches} for {Deepfake} {Detection},'' in \emph{Proc. Adv. {Neural} {Inf.} {Process.} {Syst.}}, vol.~35, 2022, pp. 4517--4530.

\bibitem{dosovitskiy_image_2021}
A.~Dosovitskiy and et~al., ``An {Image} is {Worth} 16x16 {Words}: {Transformers} for {Image} {Recognition} at {Scale},'' in \emph{International {Conf.} {Learning} {Represent.}}, 2021.

\bibitem{wang_altfreezing_2023}
Z.~Wang, J.~Bao, W.~Zhou, W.~Wang, and H.~Li, ``{AltFreezing} for {More} {General} {Video} {Face} {Forgery} {Detection},'' in \emph{Proc. {IEEE}/{CVF} {Conf.} {Comput.} {Vis.} {Pattern} {Recognit.}}, 2023, pp. 4129--4138.

\bibitem{farbman_diffusion_2010}
Z.~Farbman, R.~Fattal, and D.~Lischinski, ``Diffusion {Maps} for {Edge}-{Aware} {Image} {Editing},'' \emph{ACM Trans. Graph.}, vol.~29, no.~6, Dec. 2010.

\bibitem{sun_neural_2019}
J.~Sun and Z.~Xu, ``Neural {Diffusion} {Distance} for {Image} {Segmentation},'' in \emph{Proc. Adv. {Neural} {Inf.} {Process.} {Syst.}}, vol.~32, 2019.

\bibitem{zhao_multi-attentional_2021}
H.~Zhao, W.~Zhou, D.~Chen, T.~Wei, W.~Zhang, and N.~Yu, ``Multi-{Attentional} {Deepfake} {Detection},'' in \emph{Proc. {IEEE}/{CVF} {Conf.} {Comput.} {Vis.} {Pattern} {Recognit}}, 2021, pp. 2185--2194.

\bibitem{zhang_patch_2022}
B.~Zhang, S.~Li, G.~Feng, Z.~Qian, and X.~Zhang, ``Patch diffusion: a general module for face manipulation detection,'' in \emph{Proc. AAAI Conf. Artif. Intell.}, 2022, pp. 3243--3251.

\bibitem{chen_local_2021}
S.~Chen, T.~Yao, Y.~Chen, S.~Ding, J.~Li, and R.~Ji, ``Local {Relation} {Learning} for {Face} {Forgery} {Detection},'' 2021, pp. 1081--1088.

\bibitem{liu_spatial-phase_2021}
H.~Liu and et~al., ``Spatial-{Phase} {Shallow} {Learning}: {Rethinking} {Face} {Forgery} {Detection} in {Frequency} {Domain},'' in \emph{Proc. {IEEE}/{CVF} {Conf.} {Comput} {Vis.} {Pattern} {Recognit.}}, 2021, pp. 772--781.

\bibitem{li_frequency-aware_2021}
J.~Li, H.~Xie, J.~Li, Z.~Wang, and Y.~Zhang, ``Frequency-{Aware} {Discriminative} {Feature} {Learning} {Supervised} by {Single}-{Center} {Loss} for {Face} {Forgery} {Detection},'' in \emph{Proc. {IEEE}/{CVF} {Conf.} {Comput.} {Vis.} {Pattern} {Recognit.}}, 2021, pp. 6458--6467.

\bibitem{yu_detection_2022}
\BIBentryALTinterwordspacing
Y.~Yu, R.~Ni, W.~Li, and Y.~Zhao, ``\BIBforeignlanguage{en}{Detection of {AI}-{Manipulated} {Fake} {Faces} via {Mining} {Generalized} {Features}},'' \emph{\BIBforeignlanguage{en}{ACM Trans. Multimedia Comput., Commun., and Appl.}}, vol.~18, no.~4, pp. 1--23, 2022. [Online]. Available: \url{https://dl.acm.org/doi/10.1145/3499026}
\BIBentrySTDinterwordspacing

\bibitem{li_face_2020}
L.~Li and et~al., ``Face {X}-{Ray} for {More} {General} {Face} {Forgery} {Detection},'' in \emph{Proc. {IEEE}/{CVF} {Conf.} {Comput.} {Vis.} {Pattern} {Recognit.}}, 2020, pp. 5001--5010.

\bibitem{zhao_learning_2021}
T.~Zhao, X.~Xu, M.~Xu, H.~Ding, Y.~Xiong, and W.~Xia, ``Learning {Self}-{Consistency} for {Deepfake} {Detection},'' in \emph{Proc. {IEEE}/{CVF} {Int.} {Conf.} {Comput.} {Vis.}}, 2021, pp. 15\,023--15\,033.

\bibitem{shiohara_detecting_2022}
K.~Shiohara and T.~Yamasaki, ``Detecting {Deepfakes} {With} {Self}-{Blended} {Images},'' in \emph{Proc. {IEEE}/{CVF} {Conf.} {Comput.} {Vis.} {Pattern} {Recognit.}}, 2022, pp. 18\,720--18\,729.

\bibitem{zheng_exploring_2021}
Y.~Zheng, J.~Bao, D.~Chen, M.~Zeng, and F.~Wen, ``Exploring {Temporal} {Coherence} for {More} {General} {Video} {Face} {Forgery} {Detection},'' in \emph{Proc. {IEEE}/{CVF} {Int. {Conf.} on {Comput.} {Vis.}}}, 2021, pp. 15\,044--15\,054.

\bibitem{haliassos_lips_2021}
A.~Haliassos, K.~Vougioukas, S.~Petridis, and M.~Pantic, ``Lips {Don}'t {Lie}: {A} {Generalisable} and {Robust} {Approach} {To} {Face} {Forgery} {Detection},'' in \emph{Proc. {IEEE}/{CVF} {Conf.} {Comput.} {Vis.} {Pattern} {Recognit.}}, 2021, pp. 5039--5049.

\bibitem{khan_video_2021}
S.~A. Khan and H.~Dai, ``Video {Transformer} for {Deepfake} {Detection} with {Incremental} {Learning},'' in \emph{Proc. 29th {ACM} {Int.} {Conf.} {Multimedia}}, 2021, pp. 1821--1828.

\bibitem{sun_improving_2021}
Z.~Sun, Y.~Han, Z.~Hua, N.~Ruan, and W.~Jia, ``Improving the {Efficiency} and {Robustness} of {Deepfakes} {Detection} {Through} {Precise} {Geometric} {Features},'' in \emph{Proc. {IEEE}/{CVF} {Conf.} {Comput.} {Vis.} {Pattern} {Recognit.}}, 2021, pp. 3609--3618.

\bibitem{li_ictu_2018}
Y.~Li, M.-C. Chang, and S.~Lyu, ``In {Ictu} {Oculi}: {Exposing} {AI} {Created} {Fake} {Videos} by {Detecting} {Eye} {Blinking},'' in \emph{{IEEE} {Int. {Workshop} {Inf.} {Forensics} and {Secur.}}}, 2018, pp. 1--7.

\bibitem{karpathy_large-scale_2014}
A.~Karpathy, G.~Toderici, S.~Shetty, T.~Leung, R.~Sukthankar, and L.~Fei-Fei, ``Large-scale {Video} {Classification} with {Convolutional} {Neural} {Networks},'' in \emph{Proc. {IEEE} {Conf.} {Comput.} {Vis.} {Pattern} {Recognit.}}, 2014, pp. 1725--1732.

\bibitem{simonyan_two-stream_2014}
K.~Simonyan and A.~Zisserman, ``Two-{Stream} {Convolutional} {Networks} for {Action} {Recognition} in {Videos},'' in \emph{Proc. Adv. {Neural} {Inf.} {Process.} {Syst.}}, vol.~27, 2014.

\bibitem{bertasius_is_2021}
G.~Bertasius, H.~Wang, and L.~Torresani, ``Is {Space}-{Time} {Attention} {All} {You} {Need} for {Video} {Understanding}?'' in \emph{Proc. 38th {Int.} {Conf.} {Machi.} {Learn.}}, vol. 139, 2021, pp. 813--824.

\bibitem{arnab_vivit_2021}
A.~Arnab, M.~Dehghani, G.~Heigold, C.~Sun, M.~Lučić, and C.~Schmid, ``{ViViT}: {A} {Video} {Vision} {Transformer},'' in \emph{Proc. {IEEE}/{CVF} {Int.} {Conf.} {Comput.} {Vis.}}, 2021, pp. 6836--6846.

\bibitem{zhao_istvt_2023}
C.~Zhao, C.~Wang, G.~Hu, H.~Chen, C.~Liu, and J.~Tang, ``{ISTVT}: {Interpretable} {Spatial}-{Temporal} {Video} {Transformer} for {Deepfake} {Detection},'' \emph{IEEE Trans. Inf. Forensics Security}, vol.~18, pp. 1335--1348, 2023.

\bibitem{coifman_diffusion_2006}
R.~R. Coifman and S.~Lafon, ``Diffusion maps,'' \emph{Appli. Comput. harmonic analysis}, vol.~21, no.~1, pp. 5--30, 2006.

\bibitem{NIPS2017_ema}
A.~Tarvainen and H.~Valpola, ``Mean teachers are better role models: Weight-averaged consistency targets improve semi-supervised deep learning results,'' in \emph{Proc. Adv. {Neural} {Inf.} {Process.} {Syst.}}, vol.~30, 2017.

\bibitem{li_celeb-df_2020}
Y.~Li, X.~Yang, P.~Sun, H.~Qi, and S.~Lyu, ``Celeb-{DF}: {A} {Large}-{Scale} {Challenging} {Dataset} for {DeepFake} {Forensics},'' in \emph{Proc. {IEEE}/{CVF} {Conf.} {Comput.} {Vis.} {Pattern} {Recognit.}}, 2020, pp. 3207--3216.

\bibitem{zi_wilddeepfake_2020}
B.~Zi, M.~Chang, J.~Chen, X.~Ma, and Y.-G. Jiang, ``{WildDeepfake}: {A} {Challenging} {Real}-{World} {Dataset} for {Deepfake} {Detection},'' in \emph{Proc. 28th {ACM} {Int.} {Conf.} {Multimedia}}, 2020, pp. 2382--2390.

\bibitem{dolhansky_deepfake_2019}
B.~Dolhansky, R.~Howes, B.~Pflaum, N.~Baram, and C.~C. Ferrer, ``The {Deepfake} {Detection} {Challenge} ({DFDC}) {Preview} {Dataset},'' \emph{arXiv preprint arXiv:1910.08854}, Oct. 2019.

\bibitem{dolhansky_deepfake_2020}
B.~Dolhansky and et~al., ``The deepfake detection challenge (dfdc) dataset,'' \emph{arXiv preprint arXiv:2006.07397}, 2020.

\bibitem{jiang_deeperforensics-10_2020}
L.~Jiang, R.~Li, W.~Wu, C.~Qian, and C.~C. Loy, ``{DeeperForensics}-1.0: {A} {Large}-{Scale} {Dataset} for {Real}-{World} {Face} {Forgery} {Detection},'' in \emph{Proc. {IEEE}/{CVF} {Conf.} {Comput.} {Vis.} {Pattern} {Recognit.}}, 2020, pp. 2889--2898.

\bibitem{yang_masked_2023}
Z.~Yang, J.~Liang, Y.~Xu, X.-Y. Zhang, and R.~He, ``Masked {Relation} {Learning} for {DeepFake} {Detection},'' \emph{IEEE Trans. Inf. Forensics Security}, vol.~18, pp. 1696--1708, 2023.

\bibitem{qian_thinking_2020}
Y.~Qian, G.~Yin, L.~Sheng, Z.~Chen, and J.~Shao, ``Thinking in {Frequency}: {Face} {Forgery} {Detection} by {Mining} {Frequency}-{Aware} {Clues},'' in \emph{Proc. Eur. Conf. Comput. Vis.}, 2020, pp. 86--103.

\bibitem{sun_dual_2022}
K.~Sun, T.~Yao, S.~Chen, S.~Ding, J.~Li, and R.~Ji, ``Dual {Contrastive} {Learning} for {General} {Face} {Forgery} {Detection},'' 2022, pp. 2316--2324.

\bibitem{zhuang_uia-vit_2022}
W.~Zhuang and et~al., ``{UIA}-{ViT}: {Unsupervised} {Inconsistency}-{Aware} {Method} {Based} on {Vision} {Transformer} for {Face} {Forgery} {Detection},'' in \emph{Proc. Eur. Conf. Comput. Vis.}, 2022, pp. 391--407.

\bibitem{wang_lisiam_2022}
J.~Wang, Y.~Sun, and J.~Tang, ``{LiSiam}: {Localization} {Invariance} {Siamese} {Network} for {Deepfake} {Detection},'' \emph{IEEE Trans. Inf. Forensics Security}, vol.~17, pp. 2425--2436, 2022.

\bibitem{cdin_tcsvt}
H.~Wang, Z.~Liu, and S.~Wang, ``Exploiting complementary dynamic incoherence for deepfake video detection,'' \emph{IEEE Trans. Circuits Syst. Video Technol.}, vol.~33, no.~8, pp. 4027--4040, 2023.

\bibitem{lfdnet_tcsvt}
Z.~Guo, L.~Wang, W.~Yang, G.~Yang, and K.~Li, ``Ldfnet: Lightweight dynamic fusion network for face forgery detection by integrating local artifacts and global texture information,'' \emph{IEEE Trans. Circuits Syst. Video Technol.}, vol.~34, no.~2, pp. 1255--1265, 2024.

\bibitem{cao_end--end_2022}
J.~Cao, C.~Ma, T.~Yao, S.~Chen, S.~Ding, and X.~Yang, ``End-to-{End} {Reconstruction}-{Classification} {Learning} for {Face} {Forgery} {Detection},'' in \emph{Proc. {IEEE}/{CVF} {Conf.} {Comput.} {Vis.} {Pattern} {Recognit.}}, 2022, pp. 4113--4122.

\bibitem{gu_hierarchical_2022}
Z.~Gu, T.~Yao, Y.~Chen, S.~Ding, and L.~Ma, ``Hierarchical {Contrastive} {Inconsistency} {Learning} for {Deepfake} {Video} {Detection},'' in \emph{Proc. Eur. Conf. Comput. Vis.}, 2022, pp. 596--613.

\bibitem{zhang_deepfake_2022}
D.~Zhang, F.~Lin, Y.~Hua, P.~Wang, D.~Zeng, and S.~Ge, ``Deepfake {Video} {Detection} with {Spatiotemporal} {Dropout} {Transformer},'' in \emph{Proc. 30th {ACM} {Int.} {Conf.} {Multimedia}}, 2022, pp. 5833--5841.

\bibitem{guo_exposing_2023}
Z.~Guo, G.~Yang, J.~Chen, and X.~Sun, ``Exposing {Deepfake} {Face} {Forgeries} {With} {Guided} {Residuals},'' \emph{IEEE Trans. Multimedia}, vol.~25, pp. 8458--8470, 2023.

\bibitem{yu_augmented_2023}
Y.~Yu, X.~Zhao, R.~Ni, S.~Yang, Y.~Zhao, and A.~C. Kot, ``Augmented {Multi}-{Scale} {Spatiotemporal} {Inconsistency} {Magnifier} for {Generalized} {DeepFake} {Detection},'' \emph{IEEE Trans. Multimedia}, vol.~25, pp. 8487--8498, 2023.

\bibitem{gu_delving_2022}
Z.~Gu, Y.~Chen, T.~Yao, S.~Ding, J.~Li, and L.~Ma, ``Delving into the {Local}: {Dynamic} {Inconsistency} {Learning} for {DeepFake} {Video} {Detection},'' in \emph{Proc. AAAI Conf. Artif. Intell.}, 2022, pp. 744--752.

\bibitem{chollet_xception_2017}
F.~Chollet, ``Xception: {Deep} {Learning} {With} {Depthwise} {Separable} {Convolutions},'' in \emph{Proc. {IEEE} {Conf.} {Comput.} {Vis.} {Pattern} {Recognit.}}, 2017, pp. 1251--1258.

\bibitem{hu_dynamic_2021}
Z.~Hu, H.~Xie, Y.~Wang, J.~Li, Z.~Wang, and Y.~Zhang, ``Dynamic {Inconsistency}-aware {DeepFake} {Video} {Detection}.'' in \emph{Proc. 30th Int. Joint Conf. Artif. Intell.}, 2021, pp. 736--742.

\bibitem{chai_what_2020}
L.~Chai, D.~Bau, S.-N. Lim, and P.~Isola, ``What {Makes} {Fake} {Images} {Detectable}? {Understanding} {Properties} that {Generalize},'' in \emph{Proc. Eur. Conf. Comput. Vis.}, 2020, pp. 103--120.

\bibitem{selvaraju_grad-cam_2017}
R.~R. Selvaraju, M.~Cogswell, A.~Das, R.~Vedantam, D.~Parikh, and D.~Batra, ``Grad-{CAM}: {Visual} {Explanations} {From} {Deep} {Networks} via {Gradient}-{Based} {Localization},'' in \emph{Proc. {IEEE} {Int.} {Conf.} {Comput.} {Vis.}}, 2017, pp. 618--626.

\bibitem{van2008visualizing}
L.~Van~der Maaten and G.~Hinton, ``Visualizing data using t-sne.'' \emph{J. Mach. Learn. Res.}, vol.~9, pp. 2579--2605, 2008.

\end{thebibliography}

\begin{IEEEbiography}[{\includegraphics[width=1in,height=1.25in,clip,keepaspectratio]{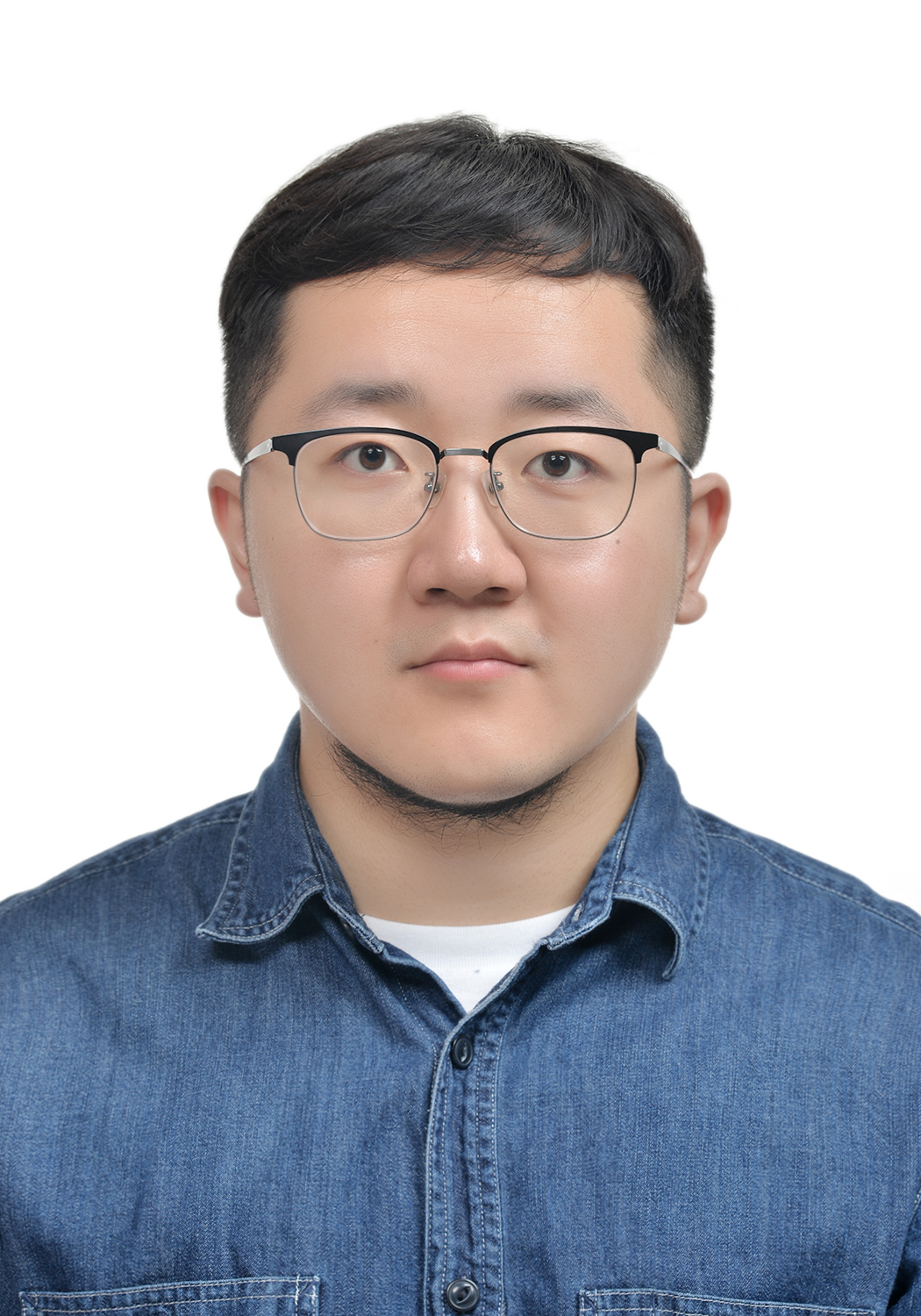}}]{Fan Nie}
received his B.S. degree and M.S. degree in the School of Computer Science from Beijing University of Posts and Telecommunications and China Electric Power Research Institute, respectively, in 2019 and 2022. He is currently pursuing Ph.D. degree in the School of Computer Science and Engineering, Sun Yat-sen University, also with Pengcheng Laboratory. His research interests include multimedia forensics, and AIGC safety.
\end{IEEEbiography}

\begin{IEEEbiography}[{\includegraphics[width=1in,height=1.25in,clip,keepaspectratio]{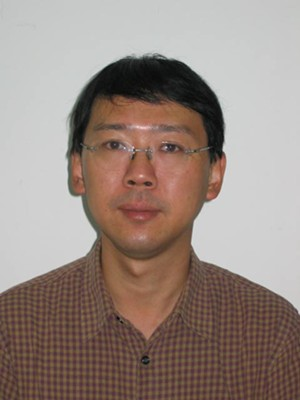}}]{Jiangqun Ni} (Member, IEEE)
received the Ph.D. degree in electronic engineering from The University of Hong Kong in 1998. Then, he was a Post-Doctoral Fellow for a joint program between Sun Yatsen University, China, and Guangdong Institute of
Telecommunication Research from 1998 to 2000. From 2001 to 2023, he was a Professor with the School of Data and Computer Science, Sun Yatsen University. He is currently a Professor with the School of Cyber Science and Technology, Sun Yat-sen University, Shenzhen, China, and also with the Department of New Networks, Peng Cheng Laboratory, Shenzhen. His research interests include data hiding, digital forensics, and image/video processing.
\end{IEEEbiography}

\begin{IEEEbiography}[{\includegraphics[width=1in,height=1.25in,clip,keepaspectratio]{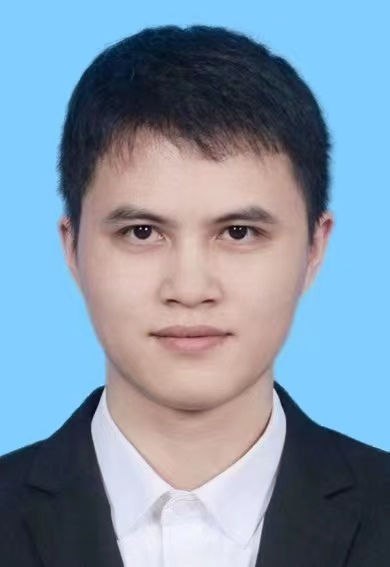}}]{Jian Zhang}
received his B.S. degree in the School of Electronics and Information Technology from Sun Yat-Sen University in 2019. He is currently pursuing Ph.D. degree in the School of Computer Science and Engineering from Sun Yat-Sen University. His research interests includes DeepFake detection and other multimedia forensics. 
\end{IEEEbiography}

\begin{IEEEbiography}[{\includegraphics[width=1in,height=1.25in,clip]{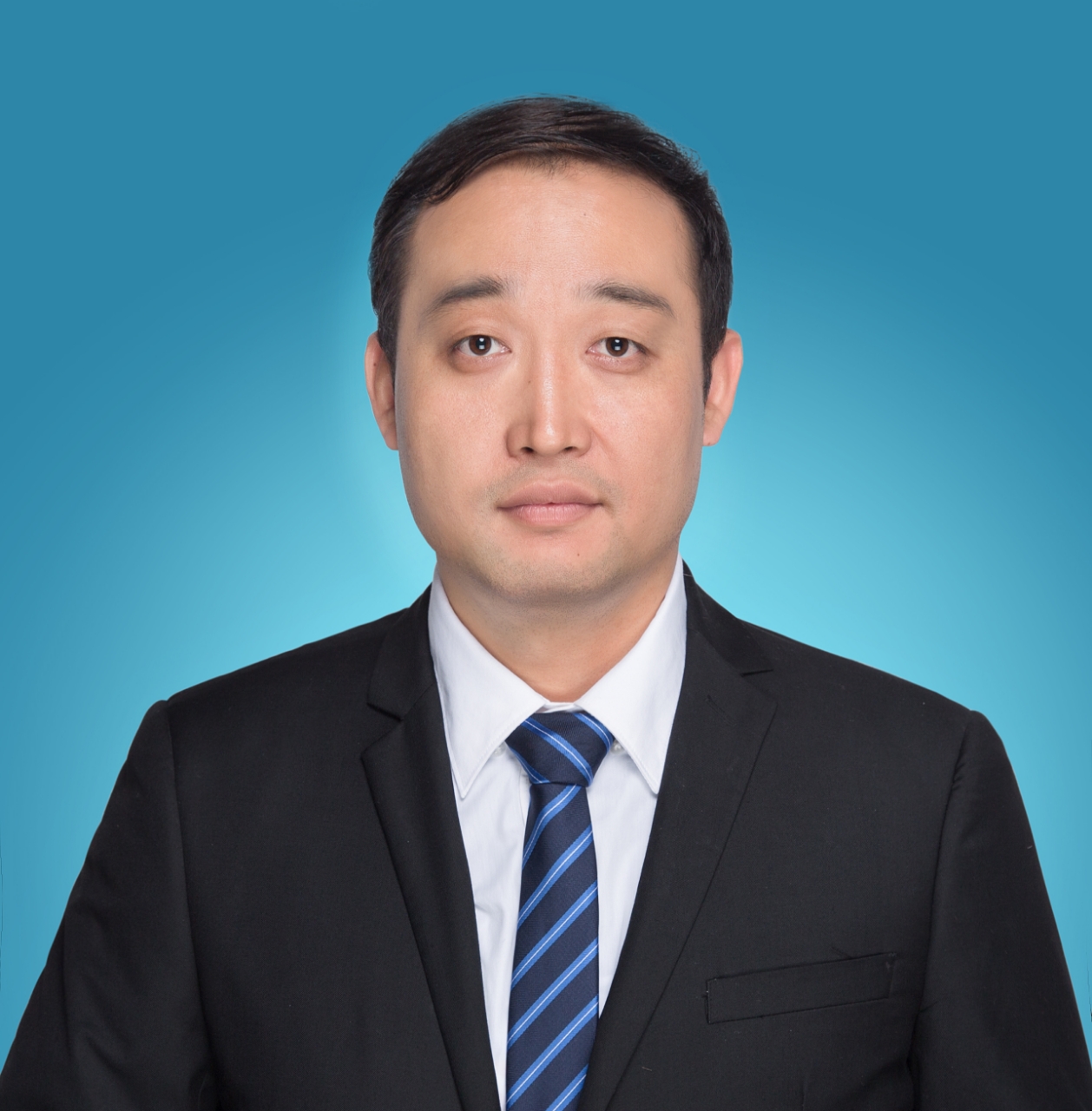}}]{Bin Zhang}
received his Ph.D. degree in the Department of Computer Science and Technology, Tsinghua University, China in 2012. He worked as a post-doctor in Nanjing Telecommunication Technology Institute from 2014 to 2017. He is now a researcher in the Department of New Networks of Pengcheng Laboratory. He publishes more than 50 papers in refereed international conferences and journals. His current research interests focus on network anomaly detection, Internet architecture and its protocols, network traffic measurement, information privacy security, etc.
\end{IEEEbiography}

\begin{IEEEbiography}[{\includegraphics[width=1in,height=1.25in,clip,keepaspectratio]{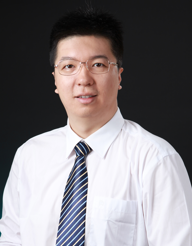}}]{Weizhe Zhang} (Senior Member, IEEE)
received the B.S, M.E and PhD degrees in computer science and technology in 1999, 2001, and 2006, respectively, from the Harbin Institute of Technology. He is currently a professor at the School of Computer Science and Technology, Harbin Institute of Technology, China, and Vice Dean of the New Network Department, Peng Cheng Laboratory, Shenzhen, China. He has authored or coauthored more than 100 academic papers in journals, books, and conference proceedings. His research interests primarily include computer network, cyberspace security,, high-performance computing, cloud computing and embedded computing.
\end{IEEEbiography}

\end{document}